\documentclass{article}

\usepackage{arxiv}

\usepackage[utf8]{inputenc} 
\usepackage[T1]{fontenc}    
\usepackage{hyperref}       
\usepackage{url}            
\usepackage{booktabs}       
\usepackage{amsfonts}       
\usepackage{nicefrac}       
\usepackage{microtype}      
\usepackage{lipsum}		
\usepackage{graphicx}
\usepackage{natbib}
\usepackage{doi}
\usepackage{float}
\usepackage{lscape} 
\usepackage{rotating}
\usepackage{tabularray}

\usepackage{lineno}
\usepackage{graphics}
\usepackage{url}
\usepackage{tabularray}
\usepackage{rotating}
\usepackage{natbib}
\usepackage{apalike}
\usepackage{amsmath,amssymb,amsfonts,amsthm}
\usepackage{multirow}
\usepackage{graphicx}
\usepackage{rotating}
\usepackage{multicol}
\usepackage{amsmath}
\usepackage{enumitem}
\usepackage{algorithmicx,algpseudocode,algorithm}
\usepackage{subfig}
\usepackage{cleveref}
\usepackage{float}
\usepackage{array}
\usepackage{algorithm}
\usepackage{caption}
\usepackage{siunitx}
\usepackage{float}
\usepackage{enumitem} 
\usepackage{xcolor}
\usepackage{graphicx}
\usepackage{color, soul, colortbl}
\usepackage{nicematrix}
\usepackage{colortbl} 
\usepackage{enumitem}
\usepackage{nameref}
\usepackage{xspace}
\usepackage{hyperref}
\usepackage{academicons}
\usepackage{orcidlink}
\usepackage{hyphenat}
\usepackage{setspace}
\usepackage{adjustbox}
\usepackage{tabularray}
\usepackage{lscape}

\title{A Comprehensive Review of Current Robot-Based Pollinators in Greenhouse Farming}

\author{ \href{https://orcid.org/0000-0000-0000-0000}{\hspace{1mm}Rajmeet  ~Singh}\\
	Khalifa University Center for Autonomous Robotic Systems (KUCARS)\\
	Khalifa University\\
	Abu Dhabi, UAE \\
	\texttt{rajmeet.bhourji@ku.ac.ae} \\
	\And
	{\hspace{1mm}lakmal ~Seneviratne} \\
	Khalifa University Center for Autonomous Robotic Systems (KUCARS)\\
	Khalifa University\\
	Abu Dhabi, UAE \\
	\texttt{lakmal.seneviratne@ku.ac.ae} \\
            \And
	{\hspace{1mm}Irfan ~Hussain} \thanks{Corresponding author} \\
	Khalifa University Center for Autonomous Robotic Systems (KUCARS)\\
	Khalifa University\\
	Abu Dhabi, UAE \\
	\texttt{irfan.hussain@ku.ac.ae} \\
}

\renewcommand{\shorttitle}{\textit{arXiv} Template}

\hypersetup{
pdftitle={A template for the arxiv style},
pdfsubject={q-bio.NC, q-bio.QM},
pdfauthor={David S.~Hippocampus, Elias D.~Striatum},
pdfkeywords={First keyword, Second keyword, More},
}

\begin{document}
\maketitle

\begin{abstract}
The decline of bee and wind-based pollination systems in greenhouses due to controlled environments and limited access has boost the importance of finding alternative pollination methods. Robotic-based pollination systems have emerged as a promising solution, ensuring adequate crop yield even in challenging pollination scenarios. This paper presents a comprehensive review of the current robotic-based pollinators employed in greenhouses.The review categorizes pollinator technologies into major categories such as air-jet, water-jet, linear actuator, ultrasonic wave, and air-liquid spray, each suitable for specific crop pollination requirements. However, these technologies are often tailored to particular crops, limiting their versatility.The advancement of science and technology has led to the integration of automated pollination technology, encompassing information technology, automatic perception, detection, control, and operation. This integration not only reduces labor costs but also fosters the ongoing progress of modern agriculture by refining technology, enhancing automation, and promoting intelligence in agricultural practices.Finally, the challenges encountered in design of pollinator are addressed, and a forward-looking perspective is taken towards future developments, aiming to contribute to the sustainable advancement of this technology.
\end{abstract}


\section{Introduction}
The agricultural sector is currently struggling with issue related to the decreases of natural pollinators, i.e honey bees, which results a significant threat to crop production. This concern is particularly pronounced within the context of greenhouse farming, where many farmers are unable to depend solely on the presence of natural pollinators within their controlled environments for efficient crop pollination.The cultivation of pollinator-dependent crops is expanding worldwide, resulting in a greater reliance on honey bees and their pollination services \citep{aizen2008long,aizen2019global}. Despite efforts to increase the honey bee colonies count globally, the growth rate has been insufficient to meet the rising demand \citep{mashilingi2022honeybees}. Consequently, this supply-demand disparity has led to pollination deficits \citep{reilly2020crop} and a subsequent surge in prices for pollination services.

Greenhouse farming has become an essential component of modern agriculture, enabling controlled and efficient cultivation of crops. However, the diminishing population of natural pollinators, including bees, presents a notable challenge to achieving effective pollination within greenhouse environments. This challenge has prompted the exploration of innovative solutions, including robot-based pollination, to ensure efficient and reliable pollination in greenhouse farming \citep{house2014fact}. There are several reasons why bees are less prevalent in greenhouses compared to outdoor environments:%
\begin{itemize}
    \item {Limited access: Greenhouses are enclosed structures designed to create a controlled environment for plant growth. The access points for bees to enter and exit may be restricted, making it difficult for them to access the greenhouse and carry out pollination effectively.}
    \item {Altered floral landscape: Greenhouse crops often consist of specific plant species that are selected for their productivity and suitability for controlled environments. This may result in a limited variety of flowers compared to natural habitats, reducing the attractiveness of the greenhouse for bees.}
    \item{Pesticide use: Greenhouses typically employ intensive pest management strategies to protect crops from pests and diseases. The use of pesticides and other chemical treatments may accidentally harm bees and deter them from entering the greenhouse.}
    \item{Temperature and humidity: Greenhouses provide a controlled climate with specific temperature and humidity conditions optimized for plant growth. These conditions may not be ideal for bees, which prefer a more natural outdoor environment.}
\end{itemize}

\begin{figure}[!ht]
\centering    {{\includegraphics[width=12cm,height=12cm,keepaspectratio]{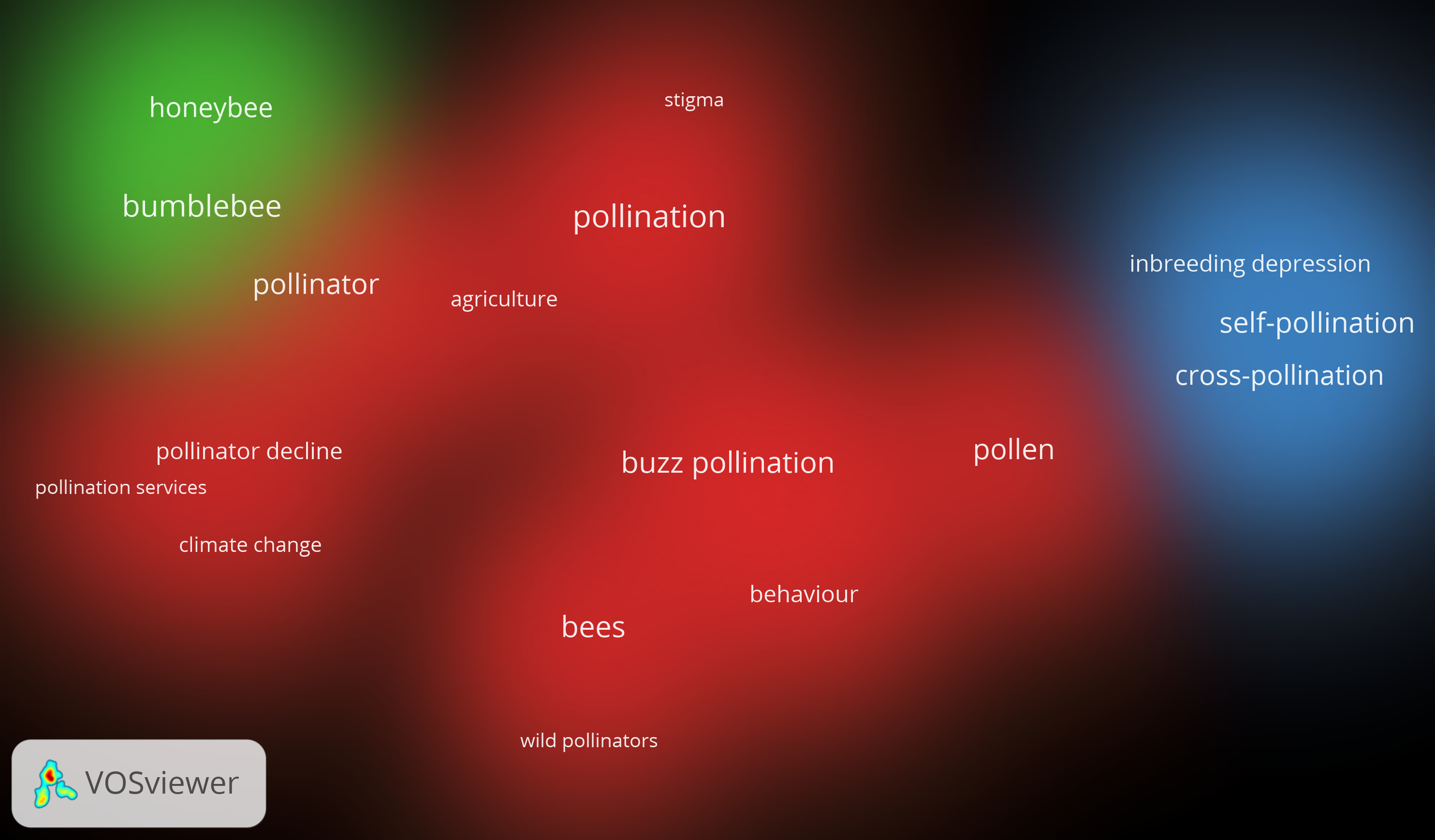} }}%
    \qquad
    \caption{Clustering density view presented the co-occurrences
of keywords that emerged from our targeted search and bibliometric analysis, shedding light on prevalent research trends in the field. Terms like ’pollination,’ ’pollinator,’ ’bees,’ ’buzz pollinator,’ and ’pollinator decline’ prominently co-occur, providing valuable insights into the core focus areas of our comprehensive review}%
    \label{fig:fig1}%
\end{figure}

As a result, alternative pollination methods are often employed to ensure successful crop pollination in these controlled environments.
Robotic precision pollination is emerging as a promising technique among the potential alternatives. By employing robotic pollinators, farmers can benefit from their consistent and predictable availability, which surpasses the variability associated with insects. Moreover, these robotic systems offer additional functionalities such as flower thinning and data gathering on crop conditions, further enhancing their value to farmers.For past years, there has been extensive research conducted on the integration of robots in the field of agriculture.Numerous notable applications include the automated harvesting of crops \citep{sarig1993robotics,van2002autonomous,baeten2008autonomous,scarfe2009development,lehnert2017autonomous}, the identification and removal of weeds \citep{lee1999robotic,blasco2002ae,aastrand2002agricultural,slaughter2008autonomous}, as well as the mapping, dataset  collection, and phenotyping \citep{weiss2011plant,cheein2011optimized,pena2013weed,mueller2017robotanist}. These advancements exemplify how robots helps to the concept of "precision agriculture", wherein crop datasets are utilized to make informed decisions regarding sustainable and tailored crop cultivation and maintenance practices \citep{bongiovanni2004precision,mcbratney2005future,zhang2012application}. In addition to their role in data gathering, robots can significantly enhance precision agriculture by executing individualized treatments and maintenance tasks. While humans may be capable of performing these tasks manually in small-scale productions, the use of robots holds great promise for large farms, where they can potentially offer increased efficiency and cost-effectiveness in the long run. In recent years, the concept of employing robots as pollinators has gained traction as a response to the decline in bee populations \citep{binns2009robotic}. The precise task of pollinating a significant number of flowers is ideally suited for robots. When we hear the term "robotic pollination", it often evokes picture of tiny, bee-like design buzzing around flowers. Researchers have been focusing on the development of tiny, bee-like flying robots \citep{wood2008robotic,ma2013controlled} and exploring control strategies for managing swarms of these miniature flyers. The ultimate goal of such research is to enable these robots to contribute to crop pollination \citep{berman2011design}. Although these early-stage demonstrations showcase the potential viability of robotic pollination, there are significant challenges that need to be addressed for flying robot-based approaches. These challenges include ensuring autonomy, extending the duration of flight, ensuring safety, and mitigating the impact of wind disturbances on the flowers \citep{wood2013flight}. Ground-based robotic pollination systems have been explored through various examples. These include a fixed crane like structure used for pollinating vanilla plants \citep{shaneyfelt2013vision}, a mobile robotic arm design intended for tall tree pollination \citep{gan2008stabilization}, and a mobile robot platform specifically designed for pollinating tomato plants \citep{yuan2016autonomous}. However, research in the field of ground-based mobile robotic pollination systems is currently limited, with only a few examples and conceptual designs available. Moreover, significant autonomy in these systems has yet to be demonstrated. This review was conducted using a targeted search approach and bibliometric analysis tools. The research focused on specific keywords such as "pollination," "pollinator," and "pollinator decline". Thorough screening was performed to meticulously evaluate the relevance of papers, specifically considering their applicability to greenhouse agricultural contexts. Papers that did not align with the research scope were excluded during this rigorous process. As a result, a carefully distilled set of 585 papers was obtained, forming the foundation of our analysis. Figure \ref{fig:fig1} illustrates the detailed bibliometric analysis conducted using VOSviewer \citep{van2010vosviewer}. This analysis allowed us to explore the intricate relationships and co-occurrences of keywords within the research efforts.
The analysis revealed the prominent occurrence of terms such as 'pollination,' 'pollinator,' 'bumblebee,' 'pollinator decline,' and 'climate change,' highlighting the prevailing research trends and focal points within the field. Furthermore, the data is presented in tabular format in Table~\ref{tab:table1}, providing a comprehensive overview of the analyzed keywords and their frequency.
\begin{table}[!ht]
\caption{Keyword Co-Occurrences and Total Link Strength of the Most Frequently Mentioned keywords}
\centering
\begin{tabular}{lcc}
\hline
\textbf{Keyword}   & \textbf{Occurrences} & \textbf{Total Link Strength} \\ \hline
pollination        & 242                  & 1301                         \\
bumblebee          & 165                  & 1221                         \\
self-pollination   & 132                  & 1120                         \\
pollinator decline & 105                  & 789                          \\
pollinator         & 80                   & 563                          \\
climate change     & 43                   & 401                          \\
buzz pollination   & 39                   & 329                          \\
vibration          & 28                   & 297                          \\ \hline
\end{tabular}
\label{tab:table1}%
\end{table}

\begin{figure}[!ht]
\centering    {{\includegraphics[width=13cm,height=13cm,keepaspectratio]{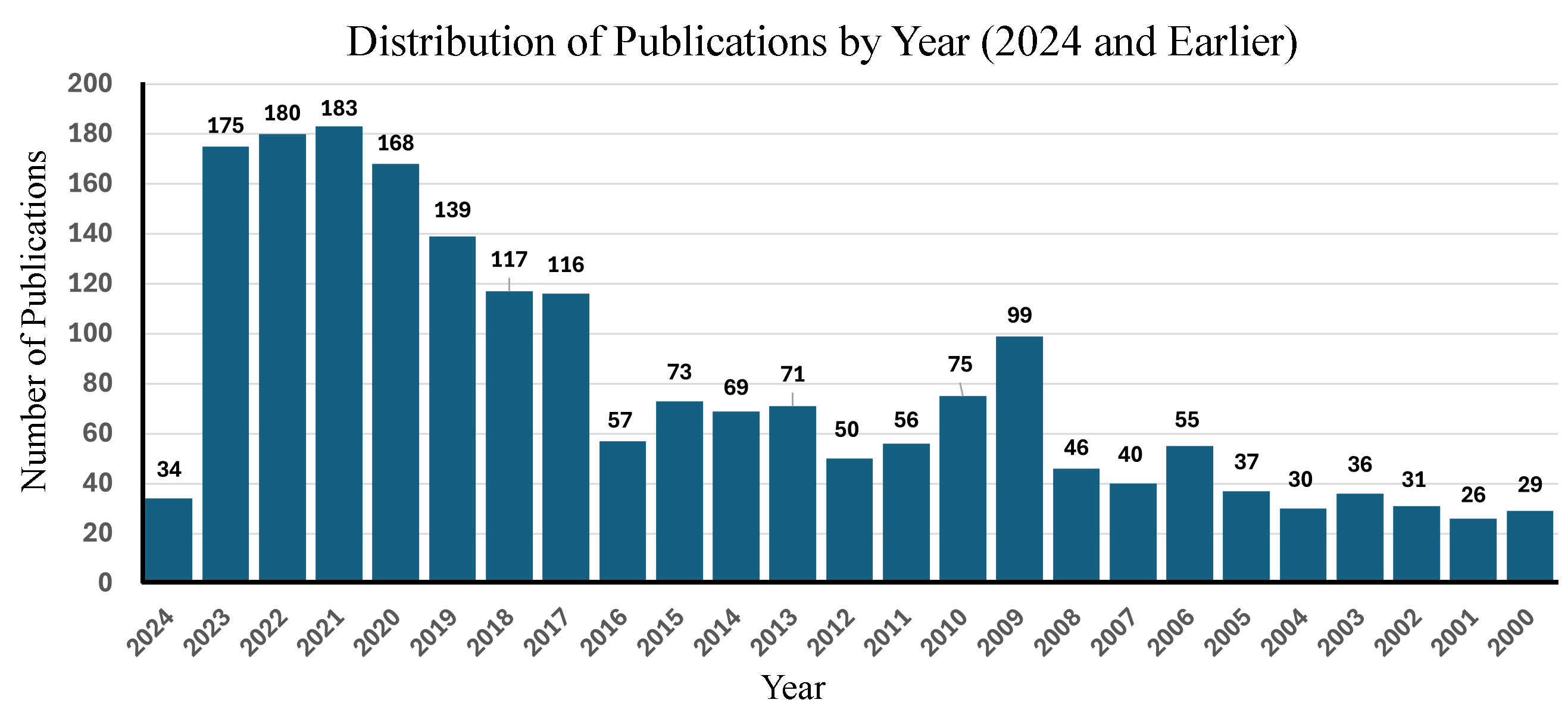} }}%
    \qquad
    \caption{Temporal Distribution of Analyzed Papers. This graph
illustrates the number of papers reviewed, plotted against their
respective publication years, offering a comprehensive view of
the research progression in our domain over time}%
    \label{fig:fig2}%
\end{figure}

\begin{figure}[!ht]
\centering    {{\includegraphics[width=13cm,height=13cm,keepaspectratio]{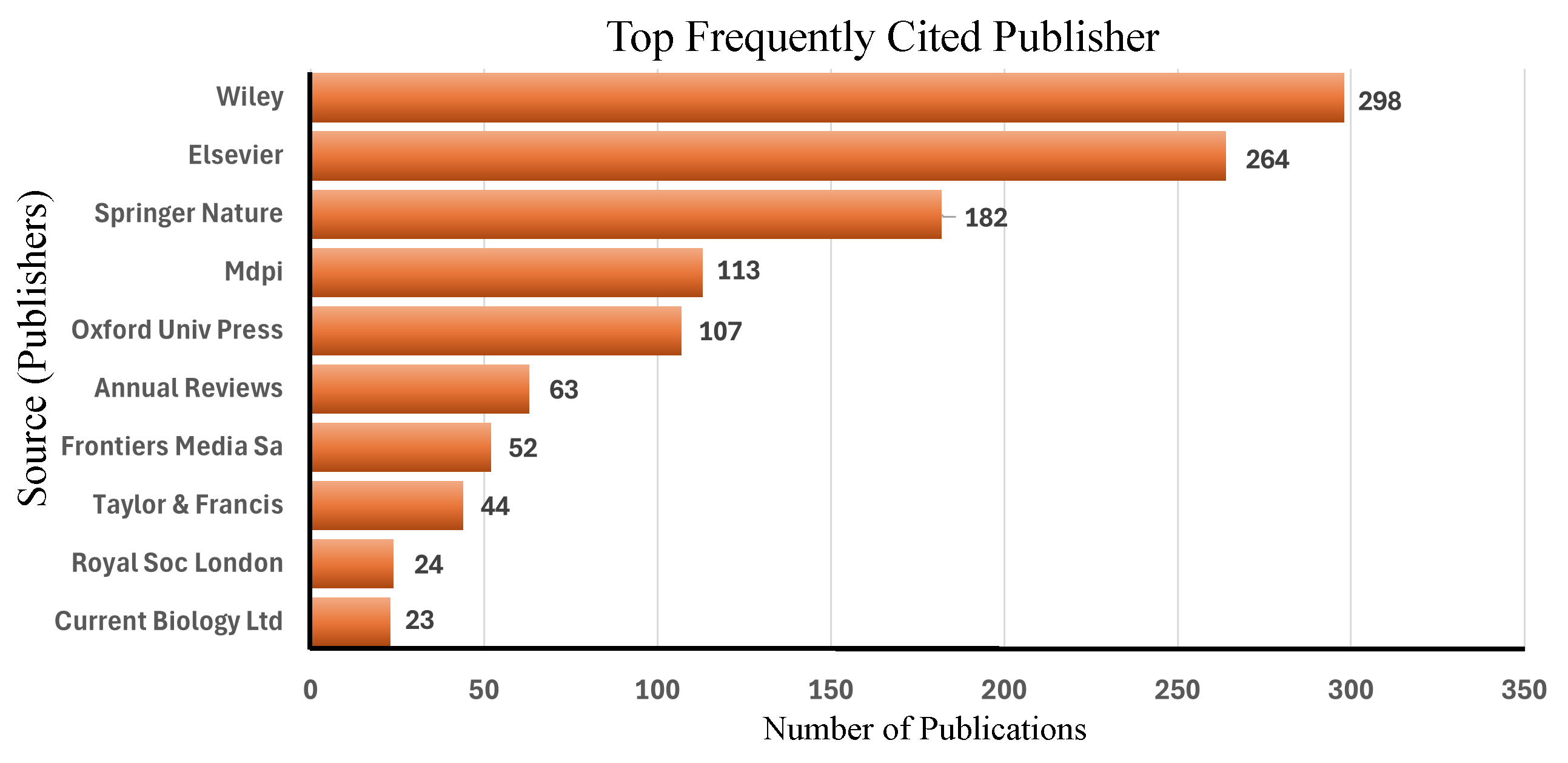} }}%
    \qquad
    \caption{Top 10 most frequently cited research publisher sources}%
    \label{fig:fig3}%
\end{figure}

\begin{figure}[!ht]
\centering    {{\includegraphics[width=12cm,height=12cm,keepaspectratio]{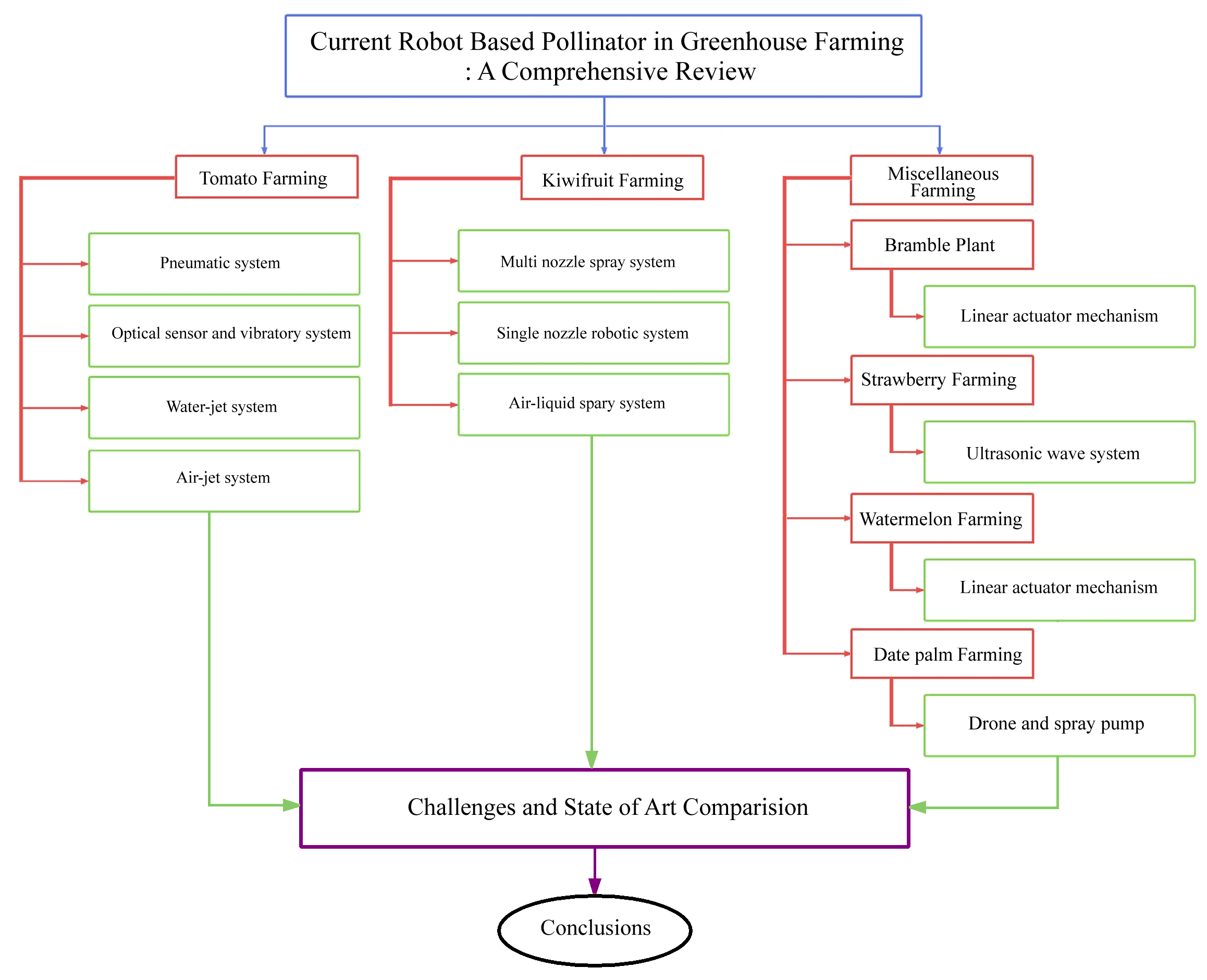} }}%
    \qquad
    \caption{Taxonomy of robot based pollinator applications in greenhouse. It illustrates the diverse approaches and technologies used in pollination for different crops }%
    \label{fig:fig4}%
\end{figure}

Figure \ref{fig:fig2} provides insights into our research data collection by illustrating the number of papers analyzed plotted against their respective publication years. This visualization offers a temporal distribution of the papers utilized in this review. In Figure \ref{fig:fig3} , we present a breakdown of the papers based on their publisher sources, delineating the number of papers obtained from different publishers. This highlights the diversity of our research sources. For a comprehensive taxonomy of the papers, please refer to Figure \ref{fig:fig4}.

The structure of the paper is as follows: In Section \ref{sec:2}, an overview of the current robot based pollinator specifically designed for tomato farming in a greenhouse environment is presented. Section \ref{sec:3} highlights the latest research on the design of robotic pollinators for Kiwi fruits. Moving forward, Section \ref{sec:4} present the current established robot based pollinator design for other crops such as strawberry, bramble plants, and watermelon farming. Furthermore, a state-of-the-art (SOTA) comparison is provided, comparing the existing design of the pollinators based on technology, crop, development phase, and gaps in Section \ref{sec:5}. Section \ref{sec:6} provides a concluding summary of the paper.

\section{Pollinators in Tomato Farming}
\label{sec:2}
\begin{figure}[!ht]
\centering
    {{\includegraphics[width=6cm,height=6cm,keepaspectratio]{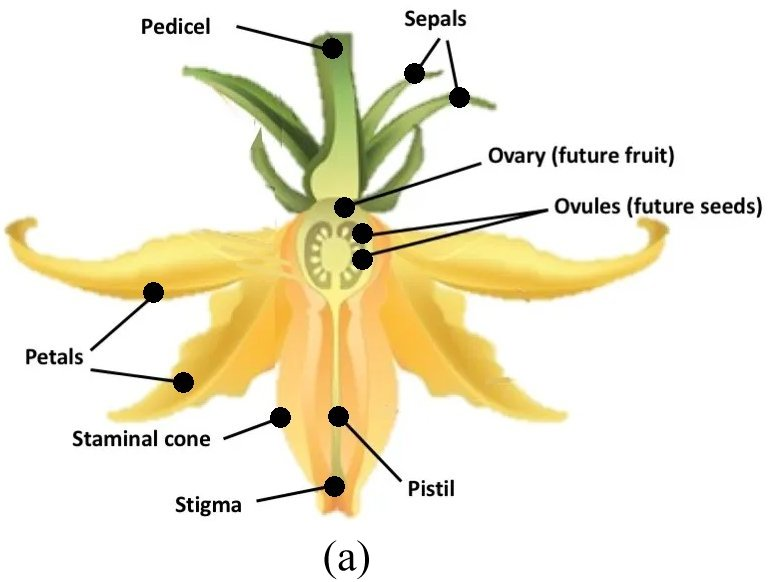} }}%
    \qquad
    {{\includegraphics[width=10cm,height=10cm,keepaspectratio]{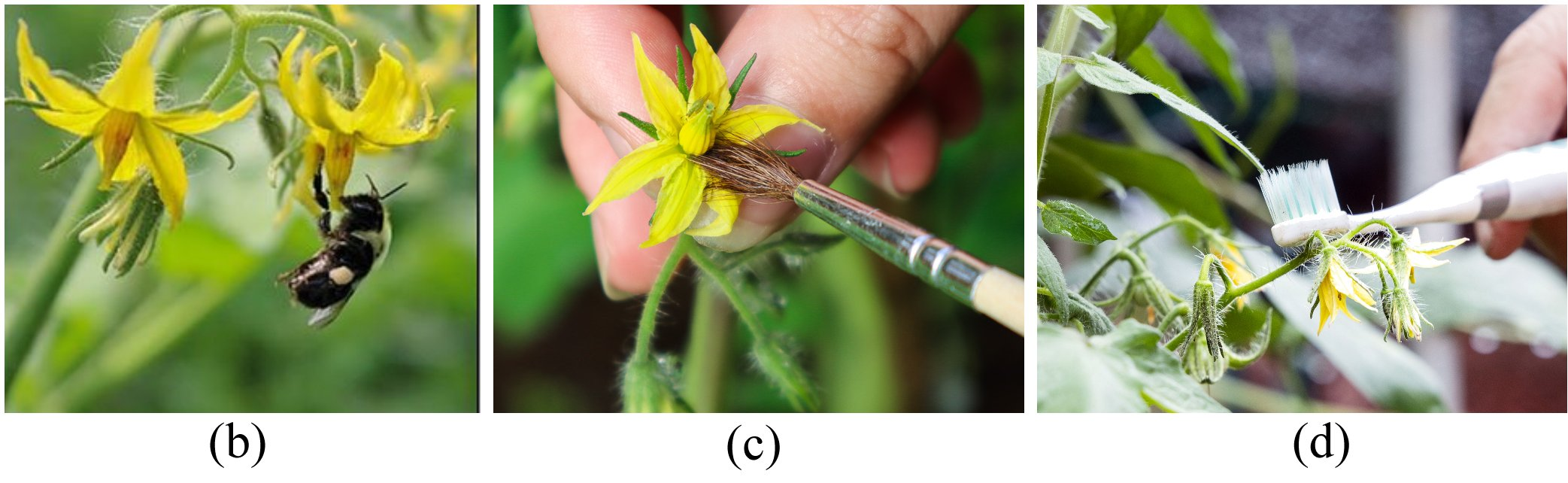} }}%
    \caption{Tomato flower pollination overview process (a) Tomato flower structure, (b) Bee pollination, (c) Hand brush pollination, and (d) Vibrator toothbrush pollination. }%
    \label{fig:fig5}%
\end{figure}

Tomato is a widely cultivated crop, occupying the top position in global crop production \citep{dingley2022precision}. As per \citep{dingley2022precision} the world wide annual production of tomato's were 189.1 million tons, strawberries (9.2 million tons), and kiwifruits (4.4 million tons) in 2021. Despite its popularity, tomatoes, strawberries, and kiwifruits face various cultivation challenges, including the critical issue of pollination in greenhouses. Pollination is the biological process in which pollen is transferred from the stamens to the pistils, ultimately resulting in the formation of seeds and fruits. While some plants require cross-pollination from different strains. The tomatoes plant can easily self-pollinate within a flower once pollen is produced. In greenhouse cultivation of tomatoes, three common methods are employed for pollination. These methods include insect pollination, artificial pollination through manual flower vibration, and hormonal pollination using plant growth regulators. Maintaining and managing insects for pollination can be challenging, especially considering their decreased activity in high temperatures during the summer, which results in reduced pollination efficiency. Additionally, the use of commercial bumblebees is limited in countries like Japan and Australia due to concerns about ecological risks \citep{dingley2022precision,nishimura2021effect}. Figure \ref{fig:fig5} shows the Tomato flower pollination overview process (a) Tomato flower structure, (b) bee pollination, (c) Hand brush pollination, and (d) Vibrator toothbrush pollination. Insect-mediated pollination aligns with natural processes, where tomato flowers are pollinated through the shaking action when honeybees and bumblebees gather pollen. However, managing and raising insects for pollination can be challenging, and the effectiveness of pollination decreases when bees become inactive in high temperatures. As a result, artificial pollination is often employed, involving manual flower vibration. In this process, farm workers visually identify the mature flowers for pollination based on their appearance and use a vibrating instrument to shake them. However, the classification of flowers for artificial pollination requires experienced and skilled farm workers. As a result, a significant number of skilled workers are needed, leading to increased expenses. Canada, Europe, and the USA encounter difficulties with the availability of skilled labor in the greenhouse sector.

\begin{figure}[!ht]
\centering    {{\includegraphics[width=13cm,height=13cm,keepaspectratio]{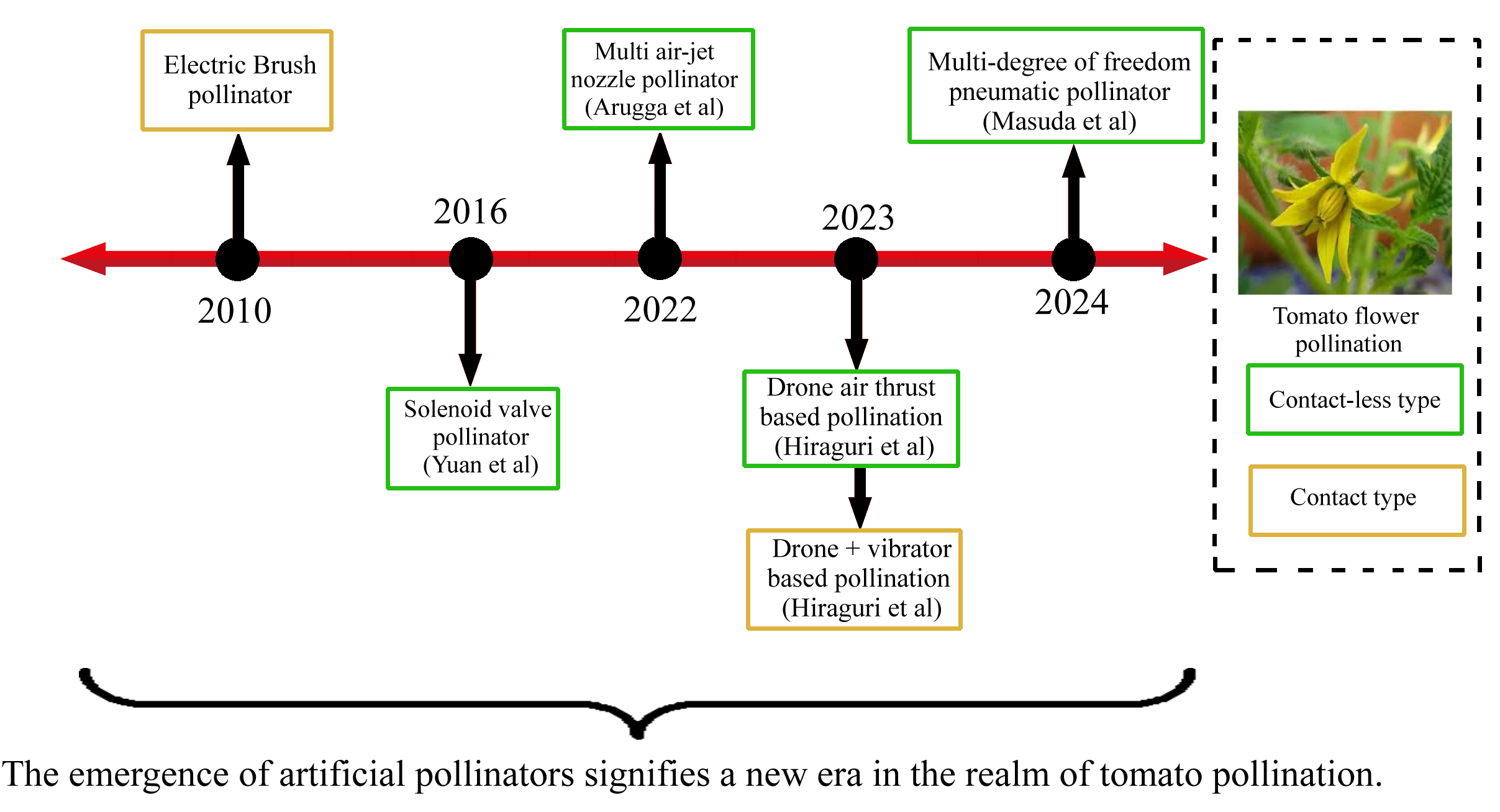} }}%
    \qquad
    \caption{Chronological over view of the most relevant pollinator methodologies for tomato farming pollination. The color of the box represents the classification based on contact or contact-less pollination. }%
    \label{fig:fig6}%
\end{figure}

Over the past decade, numerous scientist, agriculture companies, and start-ups have recognized the urgency of the pollination issue, evident from the growing number of patents related to robotic pollination devices \citep{broussard2023artificial}. 
Figure \ref{fig:fig6}  presents a timeline showcasing the evolution of artificial pollinators for tomato farming, emphasizing significant advancements in the field. Initially, tomato pollination heavily relied on natural pollinators such as honey bees and bumble bees. However, in response to the decline in bee populations, particularly within greenhouse environments, researchers have made notable progress in developing artificial pollinators specifically designed for tomato pollination. These advancements mark a significant milestone in the domain of tomato pollination, offering alternative solutions to address the challenges posed by declining bee colonies in greenhouses.

\begin{figure}[!ht]
\centering
    {\includegraphics[width=14cm,height=14cm,keepaspectratio]{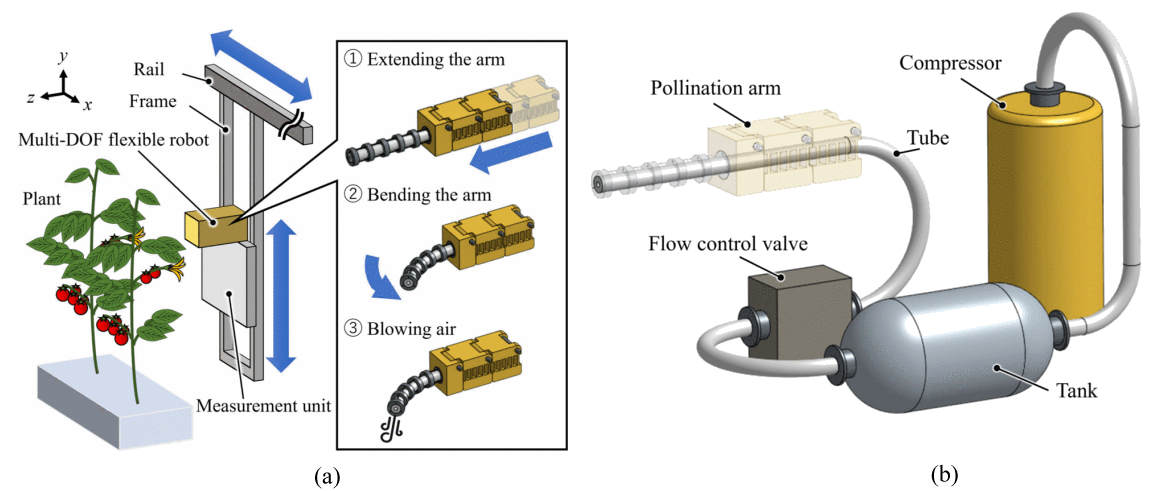} }%
    
   \caption{Multi-Degree-of-Freedom pneumatic flexible robotic pollinator. (a) A Multi-degree of freedom robotic arm, and (b) Pneumatic system for the pollination process  \citep{masuda2024development} }%
    \label{fig:fig7}%
\end{figure}

\citep{masuda2024development} proposed a multi-degrees-of-freedom (DOF) robotic pollinator capable of precise flower pollination. The robot utilizes the technique of simulating the effect of wind blowing based pollination. To accommodate cultivation restrictions, the robot incorporates a obstacle-free path generation, ensuring that the pollinator reaches the desired position without causing any damage to the crops or greenhouse structure. Compressed air is employed as the pollination method for tomato flowers. The pollination arm comprises a linear motion mechanism, a flexible link, and a pneumatic system. At the tip of the arm, a soft tube is attached, which uses compressed air to gently vibrate the flowers during the pollination process. The overall design of the multi-degree-of-freedom robotic arm is shown in Figure \ref{fig:fig7}. The proposed pollinator design is well-suited for self-pollinating crops, specifically tomato. However, this design is not viable for cross-pollination of crops like cucumber and watermelon, as it lacks the necessary bee scopa (bundles of fine hairs) for efficient pollen collection. Additionally, the pneumatic-based pollination system requires a significant amount of space during operation.     

\begin{figure}[!ht]
\centering
    {{\includegraphics[width=8cm,height=8cm,keepaspectratio]{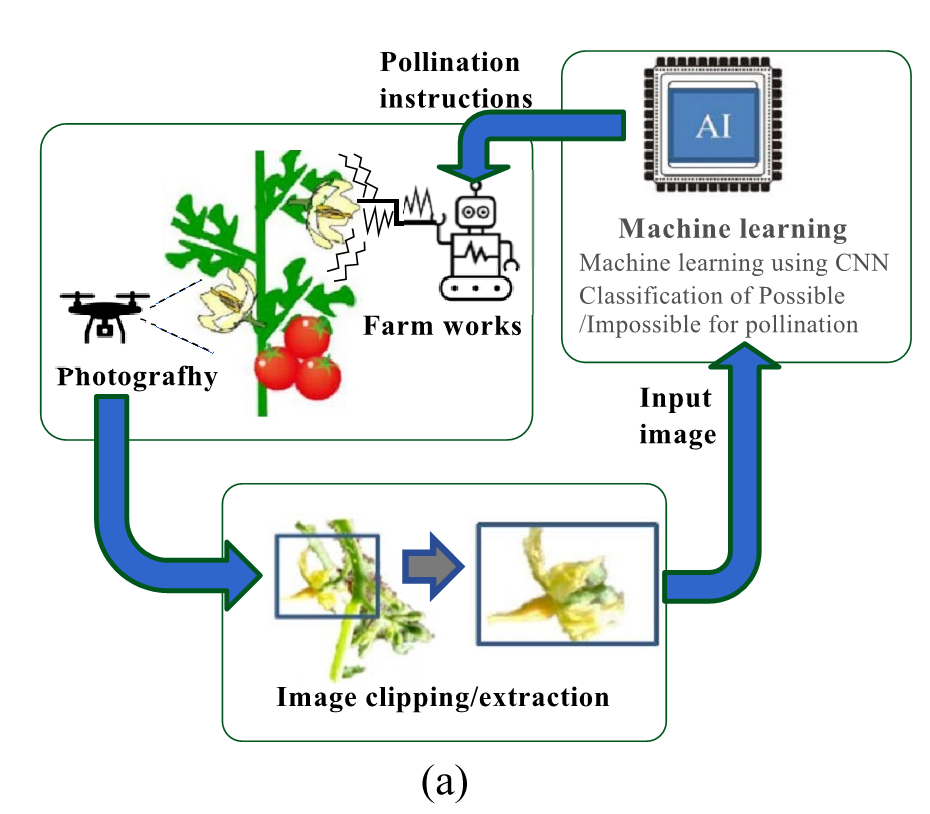} }}%
    \qquad
    {{\includegraphics[width=10cm,height=10cm,keepaspectratio]{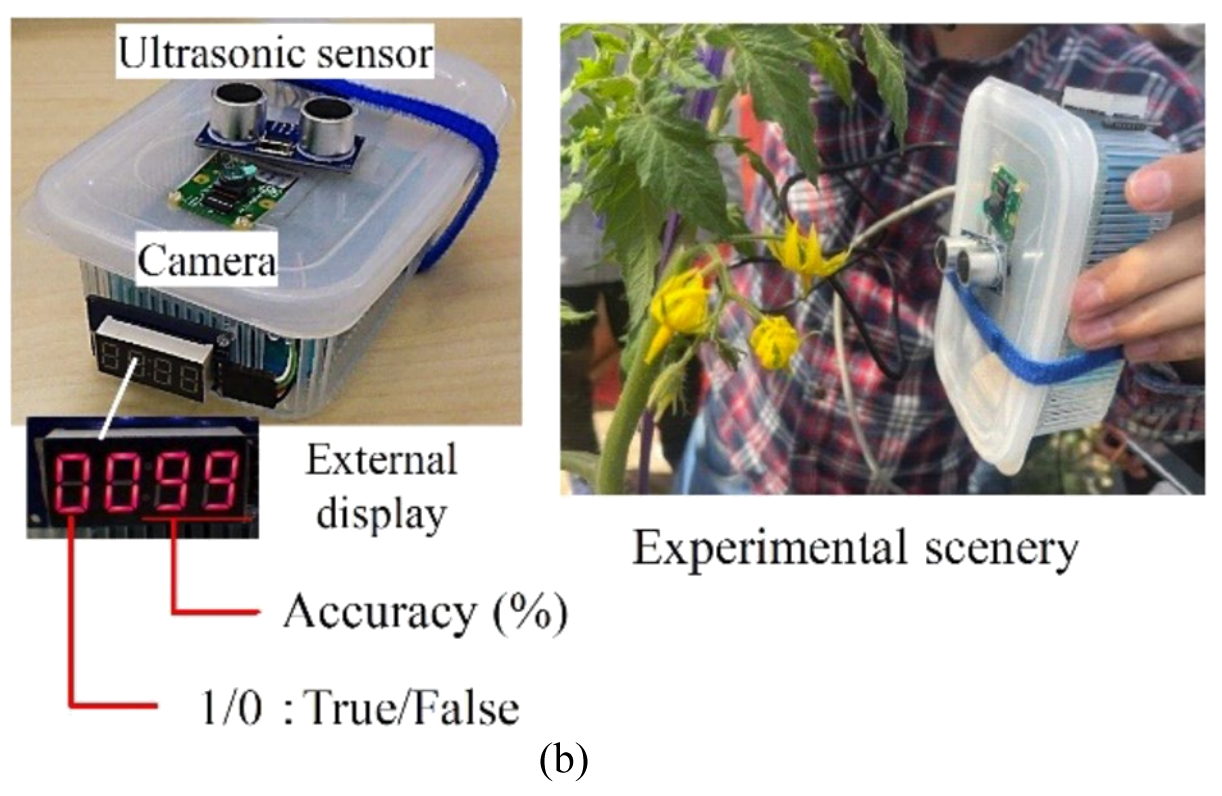} }}%
    \caption{Drone based pollination methods (a) Overall system architecture, and (b) Flower detection prototype based on ultrasonic and RGB camera  \citep{hiraguri2023shape} }%
    \label{fig:fig8}%
\end{figure}

\citep{hiraguri2023shape} proposed a approach for tomato farming involves the use of drones or robots for pollination instead of relying solely on bees and skilled labour. Team have developed a detection algorithm to identify ripe flowers, and the overall framework is illustrated in Figure \ref{fig:fig8}(a). These drones or robots need to possess similar capabilities as bees and humans to locate flowers and discern their ripeness. The primary focus is on developing technologies that enable drones and robots to autonomously distinguish flowers. In this approach, the drones cause the flowers to shake through the wind they generate. However, a limitation of drone-based pollination is the potential image blurring caused by the drone's unstable motion during image capture. To address this issue, \citep{hiraguri2023shape}developed a Gaussian filtering method to smooth the images. The developed AI model was implemented on a laboratory prototype, as depicted in Figure \ref{fig:fig8} (b), for tomato flower detection. The utilization of ultrasonic and RGB camera is employed for detection of the tomato flowers. 

\begin{figure}[!ht]
\centering
    {\includegraphics[width=14cm,height=14cm,keepaspectratio]{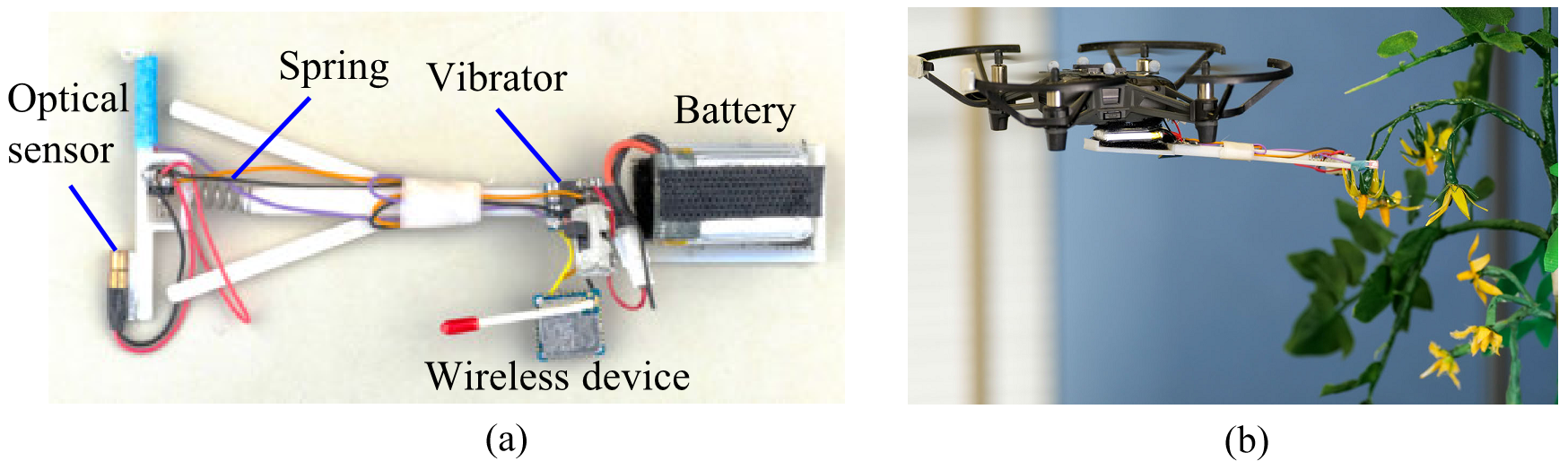} }%
    \caption{Drone and vibrator based pollinator. (a) Vibratory pollinator, and (b) Vibrator mounted on the drone for tomato flower pollination \citep{hiraguri2023autonomous} }%
    \label{fig:fig9}%
\end{figure}

Figure \ref{fig:fig9}  showcases the vibrator system for drone-based pollination in the tomato pollination process. In Figure \ref{fig:fig9} (a), the overall configuration of the pollinator design is depicted, which includes a vibrator, spring, and optical sensor. The vibrator is activated upon detecting a flower through wireless communication. The drone's motion behavior is controlled using the autonomous navigation module of the Robot Operating System (ROS) architecture. Utilizing a drone-based pollination system presents limitations in effectively targeting second layer flowers, primarily due to the presence of leaves and other plant obstacles. As a result, it becomes challenging to navigate and reach the desired flowers beyond the front layer.

\begin{figure}[!ht]
\centering
    {{\includegraphics[width=12cm,height=12cm,keepaspectratio]{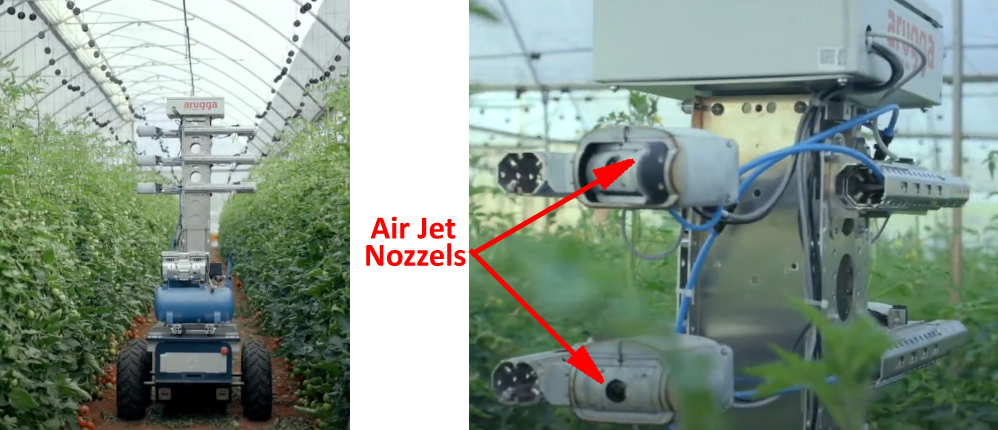} }}%
    \qquad
    \caption{Multiple Air-Jet based pollination system mounted on the mobile platform. The system is already implemented in the Israel greenhouses for commercial purpose\cite{Arugga} }%
    \label{fig:fig10}%
\end{figure}

The start-up (Arugga AI) \citep{Arugga} developed the multiple arms structure to pollinate various flowers simultaneously. Additionally, air-jets (shown in Figure \ref{fig:fig10} ) are employed to minimize the control of the actuators to position end-effectors directly onto a flower. For instance, Arugga AI has integrated sets of nozzles to facilitate the pollination of high-wire tomatoes.In the greenhouse, a mobile platform is equipped with multiple air-jet arms for efficient operation. The Arugga AI pollinator machine has entered the commercial phase and is currently being utilized in tomato greenhouse farms in Israel. This technology relies on a contactless pollination system specifically designed for tomato farming. However, the application of this technology to crops such as cucumber, watermelon, and kiwifruit has yet to be explored by Arugga AI.

\begin{figure}[!ht]
\centering    {{\includegraphics[width=12cm,height=12cm,keepaspectratio]{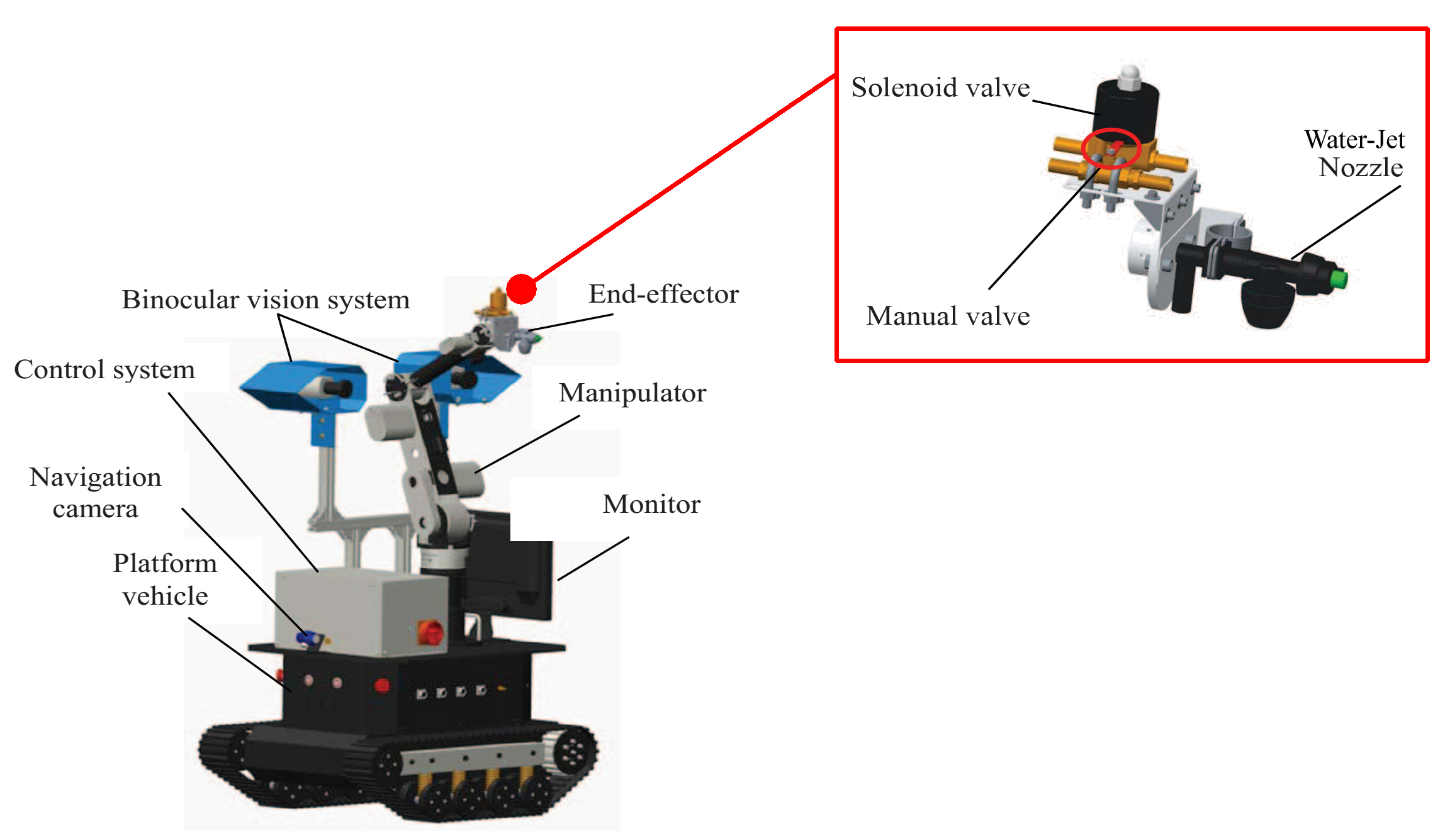} }}%
    \qquad
    \caption{Solenoid valve based water jet pollinator (end effector) is mounted on the robotic arm and navigate into greenhouse  \cite{yuan2016autonomous} }%
    \label{fig:fig11}%
\end{figure}

\citep{yuan2016autonomous} presented a dedicated end-effector designed for the task at hand. The end-effector incorporates essential components such as a spraying nozzle, a solenoid valve for controlling the spray, and a manual valve for regulating water flow. The detailed configuration is illustrated in Figure \ref{fig:fig11}. Once a female flower cluster is detected by camera, the pollinator attached to the robotic arm moves towards the target flowers. The mixture (male flower powder with water) is then sprayed onto the female flower cluster using the solenoid control. The system operates through a complex synchronization process involving multiple components. These include a camera for detection, a solenoid valve for control, and a water-jet nozzle serving as the pollinator. 

\section{Pollinators in Kiwifruits Farming}
\label{sec:3}
\begin{figure}[!ht]
\centering    {{\includegraphics[width=9cm,height=9cm,keepaspectratio]{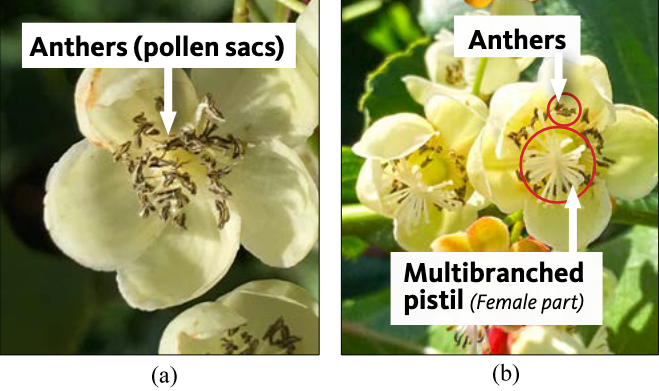} }}%
    \caption{Kiwifruit flowers (a) Male, and (b) Female. \cite{Bernadine2021} }%
    \label{fig:fig12}%
\end{figure}

Kiwi fruit, scientifically known as Actinidia deliciosa, is a popular and nutritious fruit that is widely cultivated worldwide. Native to China, this vine-like plant belongs to the family Actinidiaceae. Kiwi fruit plants are known for their distinctive appearance, with fuzzy brown skin and vibrant green flesh with tiny black seeds. The fruit is highly regarded for its unique sweet-tart flavor and rich vitamin C content. Kiwi fruit plants  have separate male and female flowers. Successful pollination is essential for fruit production, as they do not self-pollinate \citep{meroi2021fragrance,borghezan2011vitro}. 
Although male plants do not bear fruit, their role in pollination and the subsequent fruit production of female plants. In most species of kiwifruit, male and female vines can be easily separated from each other during the blooming period. Male flowers exclusively possess pollen-producing anthers, while female flowers do not produce viable pollen. The key characteristic that sets female flowers apart is their central, multi branched pistil. Following successful pollination and fertilization, this pistil develops into the fruit. While the size of the flowers may vary across different kiwifruit species, it is the presence of the pistil that allows for the differentiation between male and female flowers. The male and female kiwifruit flowers are mentioned in Figure \ref{fig:fig12}. 
 
To overcome the limitations of natural pollination and enhance the yield and quality of kiwifruit trees, artificial pollination methods are frequently employed in the breeding process. Techniques like cross-pollination, hand brush, and the use of manual or electric pollen sprayers are mostly utilized in open farming. These methods serve as effective alternatives, compensating for the inadequacies of natural pollination \citep{saez2019pollination,yang2020soap}. However, it is worth noting that these processes typically demand high skill labor \citep{chechetka2017materially,abbate2021pollination,oh2021response}

\begin{figure}[!ht]
\centering    {{\includegraphics[width=14cm,height=14cm,keepaspectratio]{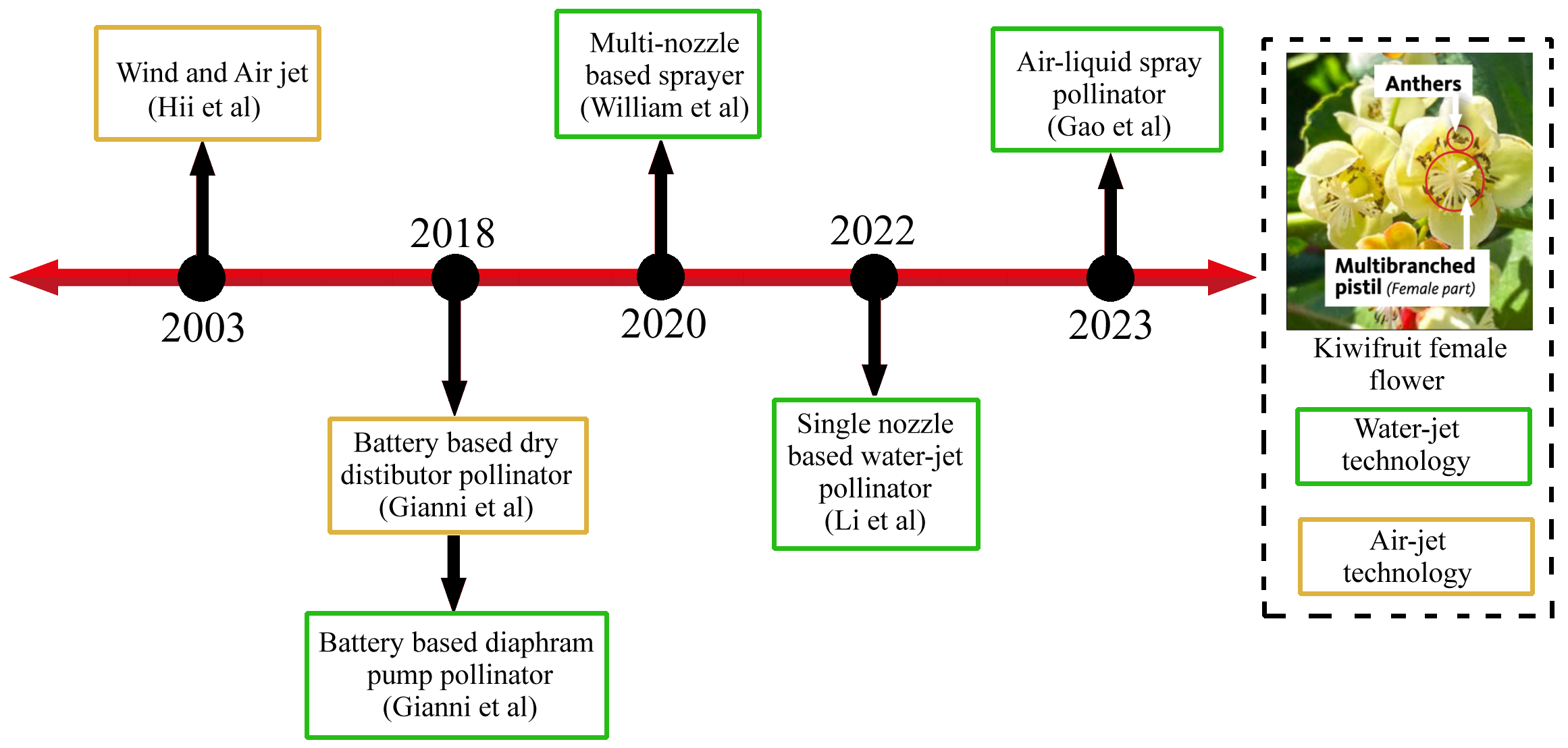} }}%
    \qquad
    \caption{Chronological overview of the artificial pollinator for kiwifruit farming. The color of the box represents the classification of pollinator based on technology i.e. air-jet or water-jet}%
    \label{fig:fig13}%
\end{figure}

Figure \ref{fig:fig13} highlights the evolution of artificial pollinators for kiwifruit farming. The researchers initially investigated the impact of wind on kiwifruit flowers to develop air-jet-based pollinators. Subsequently, water-jet-based pollinators were developed and implemented in open farms. Moreover, the researchers introduced diaphragm-based pumps, dry distributors, and air-liquid sprayers as pollination tools for kiwifruit.farming.

\begin{figure}[!ht]
\centering    {{\includegraphics[width=10cm,height=10cm,keepaspectratio]{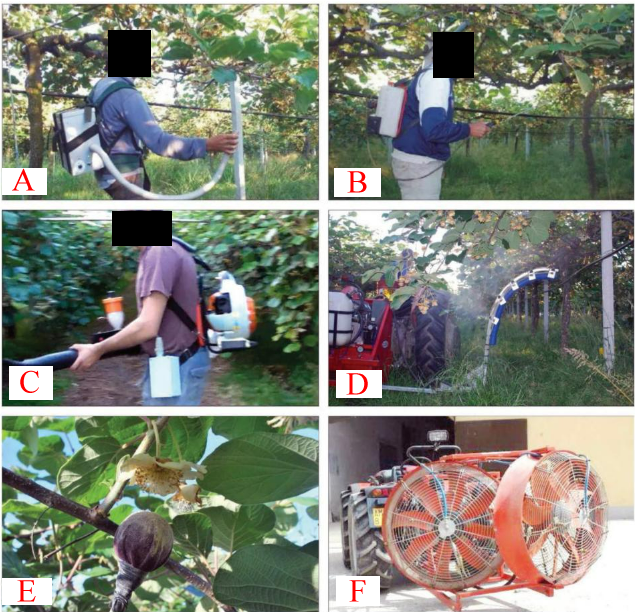} }}%
    \qquad
    \caption{The commercial available pollination system for kiwifruit for open area farming (a) Battery based dry distributor for pollen, (b) Battery based diaphragm pump for liquid pollination, (c) Engine based blower for dry pollen's, (d) Sprayer with fogger type nozzles attached to tractor, (e) Manual dry pollination, and (f) Fan attached to tractor for dry pollination\citep{gianni2018artificial} }%
    \label{fig:fig14}%
\end{figure}

\citep{hii2003modelling,hii2004kiwifruit} studied the effect of wind and air-jet on the kiwifruit flower. The computational fluid dynamics (CFD) is proposed to generate the three dimensional air flow field around the single kiwifruit flower. \citep{gianni2018artificial} presented the artificial pollinators for kiwifruit farming. The battery operated dry distributor and diaphragm pump is used for open farming of the kiwifruit. The presented artificial pollinators are suitable for open area farming.
The commercial available equipment's used in the pollination of kiwifruit in open farm is presented in Figure \ref{fig:fig14}.

\begin{figure}[!ht]
\centering    {{\includegraphics[width=13cm,height=13cm,keepaspectratio]{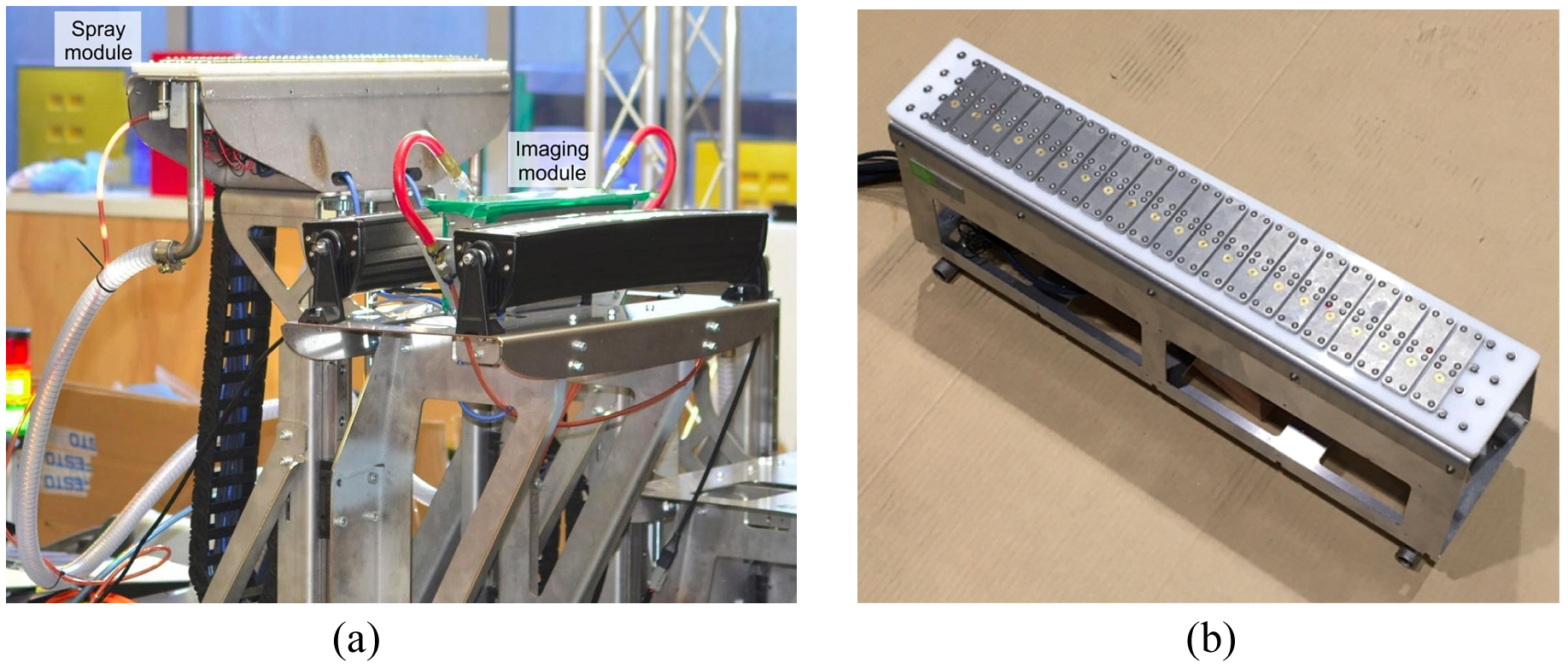} }}%
    \caption{Spray based pollinator design (a) Pollinator module mounted on the mobile platform, and (b) Spray manifold design. \citep{williams2020autonomous} }%
    \label{fig:fig15}%
\end{figure}

A design proposal has been put forth for a kiwifruit plant pollinator that incorporates multiple nozzles \citep{williams2020autonomous}. The pollination system includes a spray module equipped with 20 air and liquid nozzles, which are responsible for generating the spray. To facilitate mobility, the spray module is mounted on a mobile platform. Figure \ref{fig:fig15} (a) and (b) depicts the spray based pollinator design. The activation of a specific nozzle within the pollination system is determined by the location of the detected female flower. The control module takes the image of the detected female flower as input. However, it is important to note that the proposed pollinator has a limited action workspace.  

\begin{figure}[!ht]
\centering    {{\includegraphics[width=8cm,height=8cm,keepaspectratio]{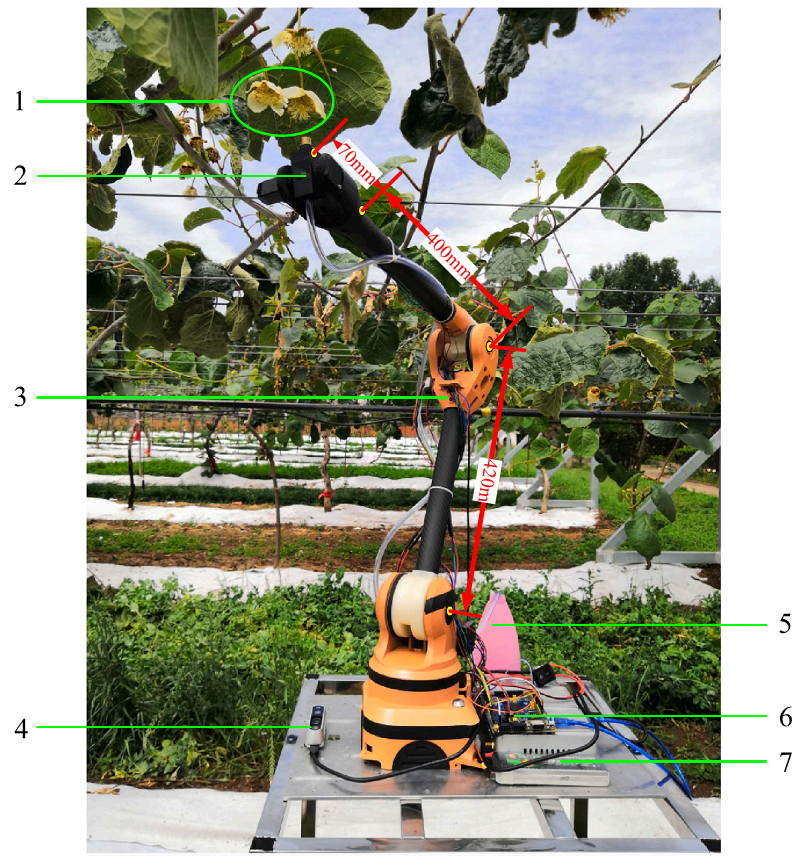} }}%
    \caption{The pollinator prototype. 1. Kiwifruit flowers, 2. Pollinator, 3. Robotic arm, 4. Binocular camera, 5. Pollen container, 6. Spray module, 7. On-board processor. \citep{li2022design} }%
    \label{fig:fig16}%
\end{figure}

\begin{figure}[!ht]
\centering    {{\includegraphics[width=12cm,height=12cm,keepaspectratio]{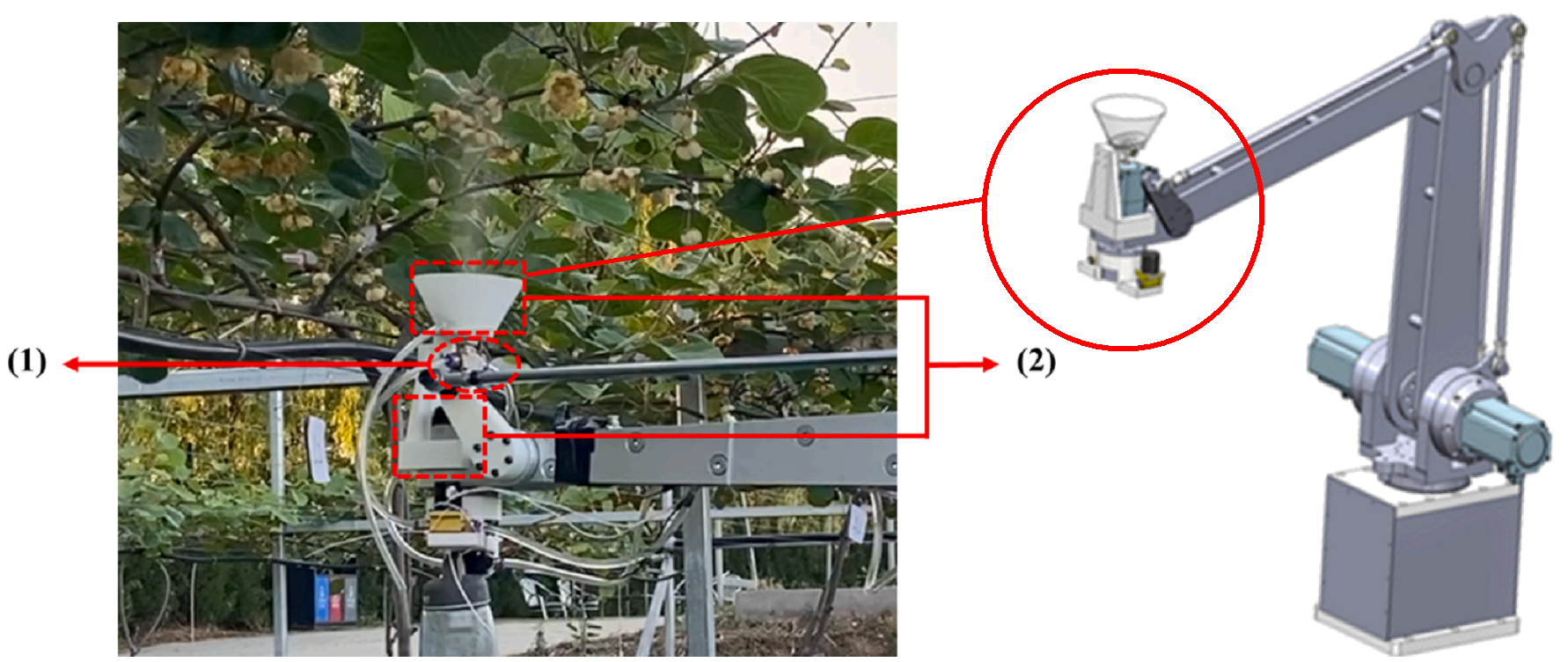} }}%
    \caption{Air–liquid spray pollinator (1) Air-liquid nozzle, and (2) Nozzle mounting brackets \citep{gao2023novel} }%
    \label{fig:fig17}%
\end{figure}

\citep{gao2023novel} and \citep{li2022design} presented a design for a kiwifruit plant pollinator that utilizes a single spray nozzle. This nozzle is installed on a lightweight robotic arm, which effectively improves the speed and agility of the robot's joints. \citep{gao2023novel} and \citep{li2022design} design surpasses \citep{williams2020autonomous} design due to the significantly greater reachability achieved by the robotic arm. The both design are shown in Figure \ref{fig:fig16} and Figure \ref{fig:fig17}. \citep{gao2023novel} developed the precise liquid pollinator for kiwifruit plants by optimizing the atomizing process parameters. To achieve precise control over the dosage of pollen suspension, the device incorporates a grating ruler to calculate the stroke of a pneumatic hydraulic cylinder. Furthermore, the system employs internal mixing air-assisted nozzles to effectively disperse the pollen suspension during the pollination process as mentioned in Figure \ref{fig:fig17}. 

\section{Pollinators for miscellaneous crops farming}
\label{sec:4}
\begin{figure}[!ht]
\centering    {{\includegraphics[width=15cm,height=15cm,keepaspectratio]{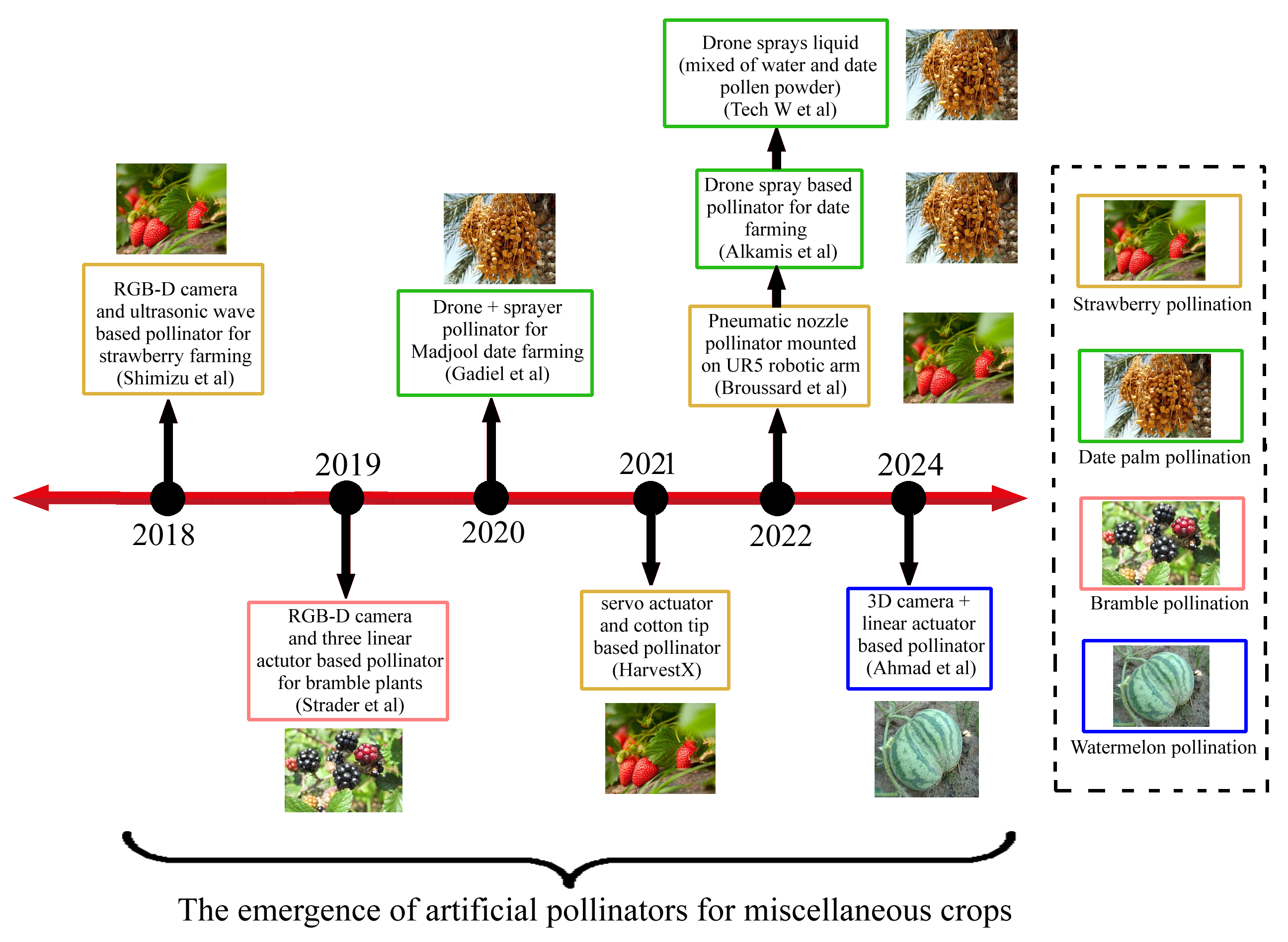} }}%
    \qquad
    \caption{Chronological overview of the artificial pollinator for strawberries, watermelons, date palms, and bramble plants. The color of the box represents the type of crops}%
    \label{fig:fig18}%
\end{figure}

In addition to tomato and kiwifruit crops, a wide range of miscellaneous crops, including strawberries, watermelons, date palms, and bramble plants, play a crucial role in global agriculture. These crops often rely on effective pollination to ensure successful fruit set and quality. However, natural pollinators, such as bees, may face challenges in adequately pollinating these crops due to factors like flower structure, bloom duration, or environmental conditions in greenhouses. To address this issue, researchers and farmers have been exploring various pollination strategies, including the use of artificial pollinators. Figure \ref{fig:fig18} highlights the evolution of artificial pollinators for strawberries, watermelons, date palms, and bramble plants with their technology implementation.

\begin{figure}[!ht]
\centering    {{\includegraphics[width=12cm,height=12cm,keepaspectratio]{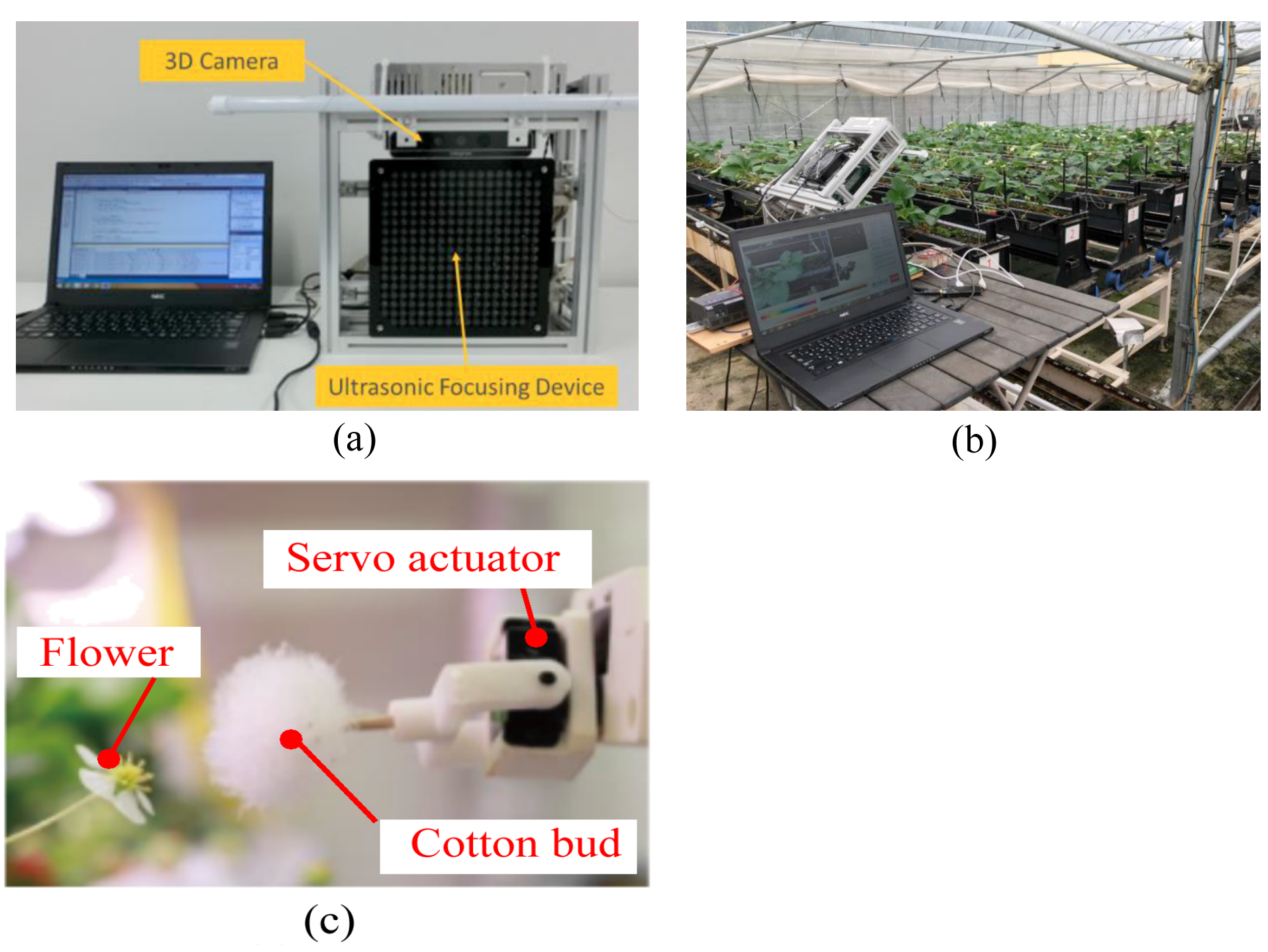} }}%
    \qquad
    \caption{Pollinator design for strawberry farming (a) Strawberry plant pollinator model using 3D camera and ultrasonic waves, and (b) Implementation in greenhouse\citep{shimizu2018development}, and (c) Servo actuator with cotton bud mounted on the pollinator tip \citep{HarvestX2021} }%
    \label{fig:fig19}%
\end{figure}

\citep{shimizu2018development} has developed a pollination system specifically designed for strawberries in a greenhouse environment. This innovative system utilizes ultrasonic radiation pressure for pollination. Flower detection is carried out using a 3D camera, while ultrasonic waves are employed to facilitate the pollination process. The effectiveness of the proposed pollination system was tested and evaluated in a strawberry farm in Japan. To cover each row in the greenhouse, an ultrasonic wave-based system is installed, leading to an increase in investment costs. 
\citep{HarvestX2021} is a startup that specializes in delivering precise and automated solutions for indoor strawberry farming. They have developed an innovative and cost-effective pollination system that incorporates a servo actuator. This system ensures high precision in the pollination process, utilizing a cotton bud mounted on the tip of the pollinator. This technology revolutionizes indoor strawberry farming by providing an efficient and reliable solution for pollination.The pneumatic gun based pollinator is proposed by \citep{broussard2023artificial,Augmentus2022} for indoor strawberry farming. The pollinator is mounted on the 6 degree of freedom Universal robot (UR5) for precise and accurate pollination of strawberry. 
Figure \ref{fig:fig19} shows the pollinator technologies for strawberry farming. (a) and (b) ultrasonic waves based pollinator for strawberry farming, \citep{shimizu2018development} and (c) Servo actuator with cotton bud mounted on the pollinator tip \citep{HarvestX2021}. 

\begin{figure}[!ht]
\centering    {{\includegraphics[width=10cm,height=10cm,keepaspectratio]{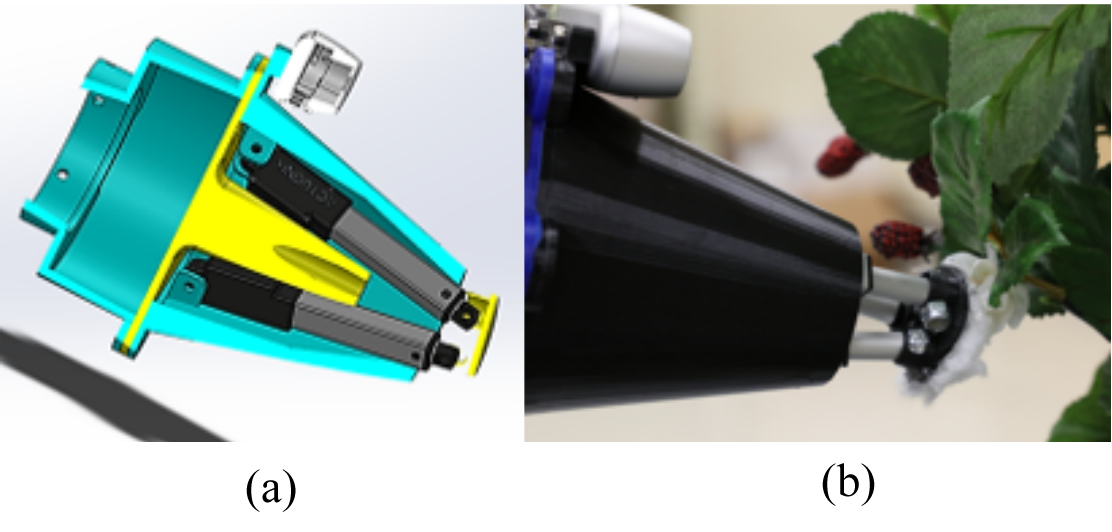} }}%
    
    \caption{Bramble plant pollinator model (a) CAD, and (b) Prototype using 3D camera and three linear actuator, the cotton bud is attached to the tip of the pollinator  \citep{strader2019flower} }%
    \label{fig:fig20}%
\end{figure}

\citep{strader2019flower} has developed a pollinator specifically designed for bramble plants, such as blackberry and raspberry, intended for greenhouse environments. The pollinator incorporates three linear servo actuators to control the movement of its tip. Additionally, an RGB-D camera is mounted along the pollinator to detect flowers and calculate the depth distance between the flower and tip. The outer body or housing of the pollinator is 3D printed using ABS material. Detailed CAD models and a prototype of the pollinator are depicted in Figure \ref{fig:fig20} (a) and (b). The proposed pollinator utilizes three linear actuators to induce vibration in the tip, which results in increased costs and complexity associated with the pollinator.

\begin{figure}[!ht]
\centering    {{\includegraphics[width=13cm,height=13cm,keepaspectratio]{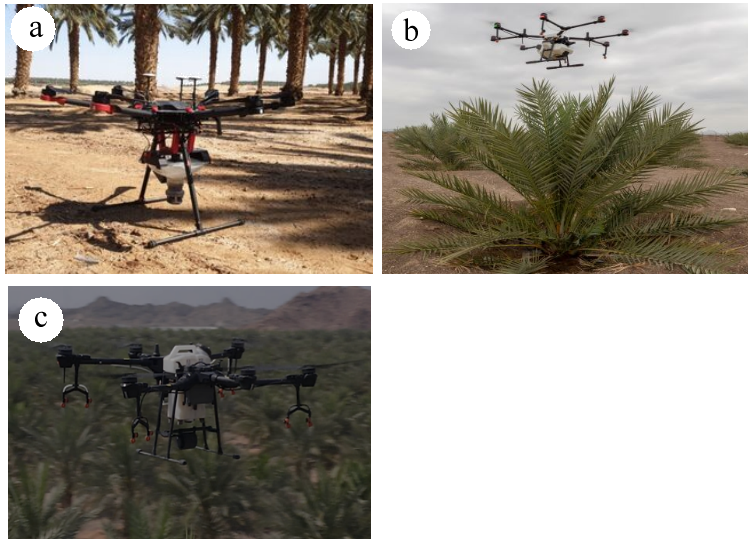} }}%
    \caption{Drone spraying pollination system for date palm farming (a) \citep{Gadiel2020}, (b) \citep{Alkhamis2022}, and (c) \citep{Wakan2022} }%
    \label{fig:fig21}%
\end{figure}

In the field of date palm pollination, \citep{Gadiel2020,Alkhamis2022,Wakan2022} have proposed the utilization of drone technology. These researchers have developed a drone equipped with a spray pump system specifically designed for date palm pollination. The drone effectively disperses male pollen, either in a water mixture or as a powder form, onto the female date palm flowers. Figure \ref{fig:fig21} visually depicts the drone with its spraying system, highlighting its application in date palm pollination.

\begin{figure}[!ht]
\centering    {{\includegraphics[width=13cm,height=13cm,keepaspectratio]{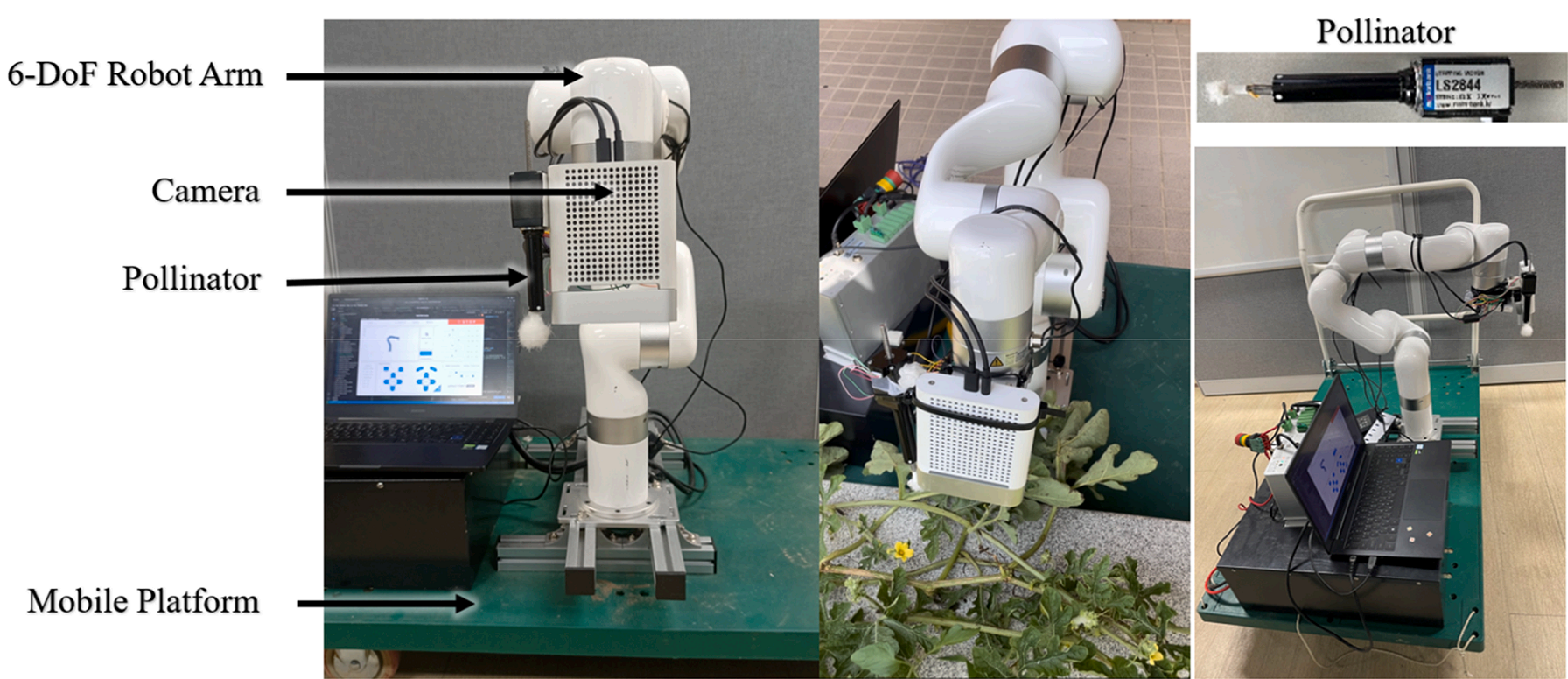} }}%
    
    \caption{Automatic watermelon flower pollinator based on 3D camera and linear actuator LS2844 \citep{ahmad2024accurate} }%
    \label{fig:fig22}%
\end{figure}

\citep{ahmad2024accurate} has proposed the robot based pollinator for watermelon farming. The 3D camera and pollinator is mounted on the 6 Degree of freedom robotic arm tip. The liner actuator LS2844 is used as pollinator. The automatic watermelon flower pollination system is depicted in Figure \ref{fig:fig22}. In the proposed system, the axis of the pollinator does not align with the axis of the camera, necessitating the development of a separate position error correction algorithm, which increases the complexity in the system.  

\section{Challenges and future trends}
\label{sec:5}
As previously mentioned, research studies have highlighted the unique advantages and promising future of pollinator equipment based on mechanization in the greenhouses. In recent years, numerous research teams have conducted extensive scientific investigations into pollination technologies, resulting in significant advancements in key areas. However, insights gained from pollination experiments have identified several challenges in current pollinator technologies, including less bees colonies, variety of crops, low efficiency, high damage rates, and limited automation. To tackle these challenges, it is crucial to develop the general and robust pollinator design which is versatile and efficient.

\subsection{Pollinator design technology}
In response to the decline in bee colonies, researchers have developed various technology-based pollinator designs for greenhouse farming. These designs aim to address the pollination needs of different crops, considering both self-pollination and cross-pollination requirements. \citep{yuan2016autonomous} designed a pollinator for tomato farming in greenhouses, utilizing a water-jet nozzle controlled by a solenoid valve. The pollinator, mounted on a mobile platform with a 6-degree-of-freedom robotic arm, offers precise control but can be complex due to multiple components. \citep{shimizu2018development} proposed an ultrasonic wave-based pollination system for self-pollination in greenhouse strawberry farming. However, this system is not suitable for cross-pollination crops like kiwifruit and dates. \citep{strader2019flower} developed a hand-in-eye pollinator for bramble plants, employing three linear actuators to vibrate the tip with a cotton bud for accurate and precise pollination. \citep{williams2020autonomous} proposed a water jet spraying-based pollinator for kiwifruit farming, activating specific nozzles based on a camera's image analysis. However, camera blurriness is a drawback due to its parallel installation with the nozzle setup. \citep{hao2023development,li2022design} improved the kiwifruit pollination design using a single water jet nozzle mounted on a 6-degree-of-freedom robotic arm, employing articulated and parallelogram mechanism-based manipulators for better flexibility. Researchers have also explored drone and air/water jet-based pollination systems for date palm farming \citep{Alkhamis2022,Gadiel2020,Wakan2022}. While autonomous navigation is still in development, skilled operators currently operate the drone for pollination. \citep{Arugga} presented a commercial prototype of a multi-air jet-based pollinator for self-pollinating tomato farming in greenhouses. \citep{hiraguri2023autonomous} developed a drone-based pollinator for tomato farming, but stability issues were encountered during vibrator operation. \citep{ahmad2024accurate} proposed a linear actuator-based pollinator design for watermelon farming, incorporating a 3D camera and a pollinator device mounted on a 6-degree-of-freedom robotic arm. However, pose estimation calculation inaccuracies between the camera and pollinator axes were addressed with a pose estimation correction algorithm, adding complexity to the system. \citep{masuda2024development} introduced a pneumatic-based pollinator design for self-pollinating tomato farming, featuring a flexible soft tube and obstacle-free path generation capability. This pollinator, mounted on a two-degree-of-freedom Cartesian system, combines tip vibration and pneumatic air blow for effective pollination. Table ~\ref{tab:table2} provides a summary of the current pollinator designs, including the technology used, working descriptions, crop types, development phases, and identified drawbacks or gaps.

\begin{sidewaystable}
\centering
\caption{Current Greenhouse Pollinators Design (State of art comparison)}
\begin{tabular}{>{\centering\hspace{0pt}}m{0.115\linewidth}|>{\centering\hspace{0pt}}m{0.098\linewidth}|>{\centering\hspace{0pt}}m{0.235\linewidth}|>{\centering\hspace{0pt}}m{0.138\linewidth}|>{\centering\hspace{0pt}}m{0.063\linewidth}|>{\centering\arraybackslash\hspace{0pt}}m{0.283\linewidth}} 
\hline
\textbf{Reference} & \textbf{Technology} & \textbf{Description} & \textbf{~Crop} & \textbf{Development} & \textbf{Drawback} \\ 
\hline
Masuda (2024) & Pneumatic tomato pollinator & pollination arm uses compressed air to gently shake flowers for effective pollen transfer during pollination.~ ~ & Tomato & prototype & cross-pollination is not feasible as bee scopa like structure is not presented \\ 
\hline
Ahmad (2024) & Linear dual shaft actuator~ ~ & 3D camera and dual shaft actuator based pollination system & Watermelon & prototype & pollinator center axis is offset from camera center axis, which needs separate pose correction algorithm~ ~ \\ 
\hline
Hiraguri(2023a) & Drone tomato pollinator~ ~ & small drones are used to shake the tomato flowers for pollination process & Tomato & prototype & target image might be blurred due to un-stability of the drone (external disturbances) \\ 
\hline
Hiraguri (2023b) & Drone and vibrator & light weight vibrator with optical sensor is mounted on the small drone for pollination~ ~~ & Tomato~ ~ & prototype & unstable during operation of vibrator. size of drone depends upon the size of the vibratory system \\ 
\hline
Hao (2023) & Air-liquid spray gun~ ~ & air-liquid atomization technique is used on 4 degree of freedom robotic arm & Kiwifruit & prototype & less workspace due to parallelogram based robotic arm structure~ ~ \\ 
\hline
Arugga (2022) & Arugga AI~ ~ & multiple targeted air jets mounted on autonomous platform~ ~ & Tomato & commercial trials & only suitable for self pollination crop (tomato) \\ 
\hline
Li (2022) & Water jet nozzle & single water jet nozzle mounted on 6 degree of freedom robotic arm & Kiwifruit & prototype & not target specific female flower as larger water-jet flow aperture  \\ 
\hline
 Alkamis (2022) Tech (2022) Gadiel (2020) & drone + spray & spray pump system is mounted on the drone & Date palm & commercial trials & operate by skilled operator, running in tele-operational mode  \\ 
\hline
Williams (2020) & Multiple Water jet nozzles~ ~ & multiple water jet nozzles mounted on the robotic platform & Kiwifruit & prototype & water spray causes camera blurriness as camera is installed parallel to the nozzle system. \\ 
\hline
Strader (2019) & Linear cylindrical actuator & three linear actuators based mechanism is used for pollination~ ~~ & Blackberry, Raspberry & prototype & three linear actuators consumed more power to make tip vibrating~ ~ \\ 
\hline
Shimizu (2018) & Ultrasonic waves & camera based ultrasonic wave generation for pollination~ ~ & Strawberry~ ~ & prototype & not feasible for cross-pollination crops \\ 
\hline
Yuan (2016) & Solenoid pollinator~ ~ & solenoid valve based water-jet pollinator & Tomato & prototype & the control of the system is challenging due to its complexity. \\ 
\hline
\label{tab:table2}%
\end{tabular}
\end{sidewaystable}

To drive future advancements in crop pollination equipment, research and development should prioritize the development of simplified structures, enhanced pollination efficiency, and reduced manufacturing costs. Current equipment utilizes various technologies such as 3D camera, linear actuators, vibrators, ultrasonic waves, air jets, water jets, and drone sprayers for the pollination process. The market for robotic pollination is still in its early growth stages, but it holds significant potential, particularly in greenhouse environments. The trend for future pollination robots is to reduce research and development costs while improving the supporting infrastructure of pollination equipment to achieve cost-effectiveness. To create intelligent, efficient, versatile, and affordable pollinators, operators should monitor the real-time operational status of each facility and optimize parameters through experimentation. Furthermore, future developments should prioritize general-purpose functionality, intelligence, and user-friendliness to fully replace manual pollination and address the bee decline issue in greenhouse farming. Currently, the development and adoption of intelligent pollinators are progressing rapidly, leading to cost reductions and improved efficiency. This trend aligns with the future direction of intelligent general-purpose pollinators. Non-contact image sensors have emerged as a crucial factor in shaping the future of pollinator design. As we look ahead, the integration of image sensors with diverse pollination technologies adds versatility and efficiency to the pollination process across different crop types in greenhouses.These advancements are driving the acceleration of the pollination process while simultaneously addressing cost-effectiveness.

\section{Conclusion}
\label{sec:6}
This paper reviews the progress of the application of pollinator design technology in the fled of crop pollination.The paper explains various technologies used for pollinator design according to type of the crops. In this paper, the self pollination and cross-pollination crops are considered i.e. tomato, strawberry, kiwifruit, watermelon, date palm, bramble plants. Robotic-based pollination has emerged as a crucial solution to supplement insect pollinators in various cropping systems. Greenhouses, with their controlled climates, pose challenges for bee survival, making robotic pollinators essential. The development of different technologies, including air-jet, water-jet, ultrasonic wave, linear actuators, and air-liquid spray, has enabled artificial pollination. Each technology has its own advantages and drawbacks, and careful consideration is needed to select the most suitable approach for specific crop requirements. This paper provides a comprehensive overview of the current state of pollinator equipment, categorizing them based on technology, crop, development phase, and identified drawbacks or gaps. This analysis helps identify areas where further research and improvement are needed to enhance pollination efficiency.With the rapid expansion of greenhouse crop plantations and rising labor costs, there is an urgent need for innovative robotic technologies to address the challenges of pollination.

In summary, this paper highlights the importance of robotic-based pollinators in greenhouse environments and presents a comprehensive analysis of existing technologies. The development of such robotic technologies will play a crucial role in ensuring sustainable and efficient crop production in the face of evolving agricultural challenges.
Non-contact image sensors have emerged as a crucial factor in shaping the future of pollinator design. Their presence is highly valuable as they facilitate the identification of target female flowers and play a significant role in recognizing mature and ready flowers for optimal pollination. These sensors offer the capability to detect and analyze flowers without physical contact, ensuring effective and precise pollination. As we look ahead, the integration of image sensors with diverse pollination technologies adds versatility and efficiency to the pollination process across different crop types in greenhouses. With ongoing technological advancements and continued government support, there are promising prospects for the research and development of intelligent general-purpose pollinator designs in the future.

\section*{Acknowledgments}
This publication is based upon work supported by the Khalifa University of Science and Technology under Grant No. RC1-2018-KUCARS-T4.

\section*{Declarations}
The authors declare that they have no known competing financial interests or personal relationships that could have appeared to influence the work reported in this paper.

\begin{itemize}
\item \textbf{Funding:}
This study was supported by the Khalifa University of Science and Technology (Grant No. RC1-2018-KUCARS-T4)
\item \textbf{Conflict of interest:}
I declare that there is no conflict of interest regarding the publication of this paper. I, corresponding author on behalf of all contributing authors, hereby declare that the information given in this disclosure is true and complete to the best of my knowledge and belief.
\item \textbf{Author contribution:}
Rajmeet Singh: Conceptualization, Methodology, Writing original draft, Visualization, Investigation. Lakmal Seneviratne: Methodology, Supervision, Project administration, Funding acquisition. Irfan Hussain: Conceptualization, Supervision, Project administration, Funding acquisition.
\end{itemize}







\end{document}